\pdfoutput=1

\documentclass[11pt]{article}

\usepackage{arabtex}

\usepackage{acl}

\usepackage{times}
\usepackage{latexsym}
\usepackage[linesnumbered,ruled,lined]{algorithm2e}
\usepackage{url}
\usepackage{latexsym}
\usepackage{amsmath,amssymb,amsfonts}
\usepackage{graphicx}
\usepackage{textcomp}
\usepackage{subcaption}
\usepackage{utf8}
\usepackage{multirow}
\usepackage{enumitem}
\usepackage{rotating}
\usepackage{colortbl}
\usepackage{tikz}
\usepackage{adjustbox}
\usepackage{soul}
\usepackage{booktabs} 
\usepackage{arydshln}
\usepackage[normalem]{ulem}
\usepackage{comment}

\usepackage{dblfloatfix}

\usepackage{enumitem}
\setlist[itemize]{align=parleft,left=5pt..12pt}

\usepackage{tikz}

\usepackage[textsize=scriptsize]{todonotes}

\usepackage[T1]{fontenc}

\usepackage[utf8]{inputenc}

\usepackage{microtype}
\usepackage{pifont}

\definecolor{lightblue}{RGB}{219,226,238}
\definecolor{lightgreen}{RGB}{217,234,211}
\definecolor{lightorange}{RGB}{255, 162, 56}
\definecolor{lightviolet}{RGB}{111, 91, 252}

\title{{\sc ReadMe++}: Benchmarking Multilingual Language Models for Multi-Domain Readability Assessment}

\author{Tarek Naous, Michael J. Ryan, Anton Lavrouk, Mohit Chandra, Wei Xu \\
  College of Computing \\
  Georgia Institute of Technology \\
  \small{
 \texttt{\{tareknaous, michaeljryan, antonlavrouk, mchandra9\}@gatech.edu; wei.xu@cc.gatech.edu}}
\\}

\begin{document}
\maketitle

\begin{abstract}

We present a comprehensive evaluation of large language models for multilingual readability assessment. Existing evaluation resources lack domain and language diversity, limiting the ability for cross-domain and cross-lingual analyses. This paper introduces {\sc ReadMe++}, a multilingual multi-domain dataset with human annotations of 9757 sentences in Arabic, English, French, Hindi, and Russian, collected from 112 different data sources. This benchmark will encourage research on developing robust multilingual readability assessment methods. Using {\sc ReadMe++}, we benchmark multilingual and monolingual language models in the supervised, unsupervised, and few-shot prompting settings. The domain and language diversity in {\sc ReadMe++} enable us to test more effective few-shot prompting, and identify shortcomings in state-of-the-art unsupervised methods. Our experiments also reveal exciting results of superior domain generalization and enhanced cross-lingual transfer capabilities by models trained on {\sc ReadMe++}. We will make our data publicly available and release a python package tool for multilingual sentence readability prediction using our trained models at: 
\url{https://github.com/tareknaous/readme}


\end{abstract}

\section{Introduction}

Readability assessment is the task of determining how difficult it is for a specific audience to read and comprehend a piece of text \cite{vajjala2022trends}. Developing methods for automatically predicting the readability of a sentence is beneficial for many applications such as controllable text simplification \cite{chi2023learning,mt}, ranking search engine results by their level of difficulty \cite{search-engine}, and selecting appropriate reading material for language learners \cite{xia2019text}. Making such technologies robust to textual variations and accessible to a global community with diverse languages requires readability prediction methods that generalize across different text domains and language families.

\begin{figure}
    \centering
    \includegraphics[width=\linewidth]{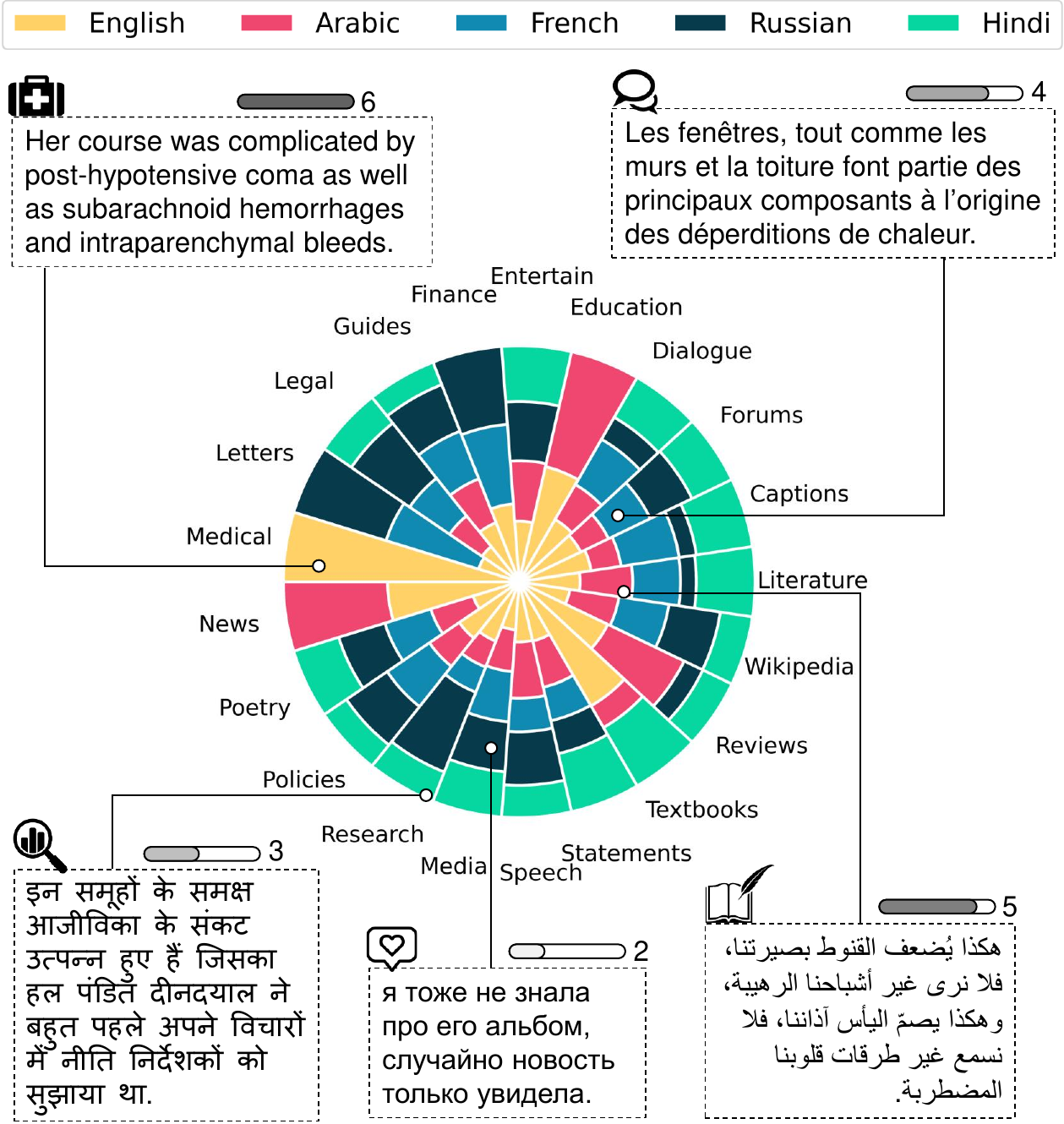}
    \caption{Language distribution per each domain in {\sc ReadMe++}. Example sentences from each language are shown along with their human-annotated readability levels on a 6-point scale (1: easiest, 6: hardest).}
    \label{fig:examples-intro}
    \vspace{-.5cm}
\end{figure}

\begin{table*}[t]
\centering
\setlength{\tabcolsep}{3pt}
\begin{adjustbox}{width=1\textwidth}
\begin{tabular}{lccl}
\toprule
\textbf{Dataset} & \textbf{Languages} & \textbf{Scripts} & \textbf{\#Data Sources}  \\ \midrule
MTDE \cite{mtde} & en, nl & Latin & 4 (Wikipedia, BNC, Dutch Parallel Corpus,   SoNaR) \\
S1131 \cite{stajner} & en & Latin & 2 (Wikipedia, Newsela) \\
CompDS \cite{brunato} & en, it & Latin & 2 (Italian UD Treebank, WSJ from Penn Treebank) \\
TextComplexityDE \cite{tcde} & de & Latin & 1 (Wikipedia, Leichte Sprache) \\
CEFR-SP \cite{cefr} & en & Latin & 3 (Wikipedia, Newsela, SCoRE) \\ \midrule
\textbf{{\sc ReadMe++} (Ours)} & ar, en, fr, hi, ru & Arabic, Brahmic, Cyrillic, Latin  & \textbf{112} (examples in Table \ref{tab:dataset-domains}; full list in Appendix \ref{appendix:domains})  \\ \bottomrule
\end{tabular}
\end{adjustbox}
\caption{Summary of readability datasets with \textit{sentence-level annotations}. Our {\sc ReadMe++} corpus provides more domain and typological diversity. There also exist more datasets with document-level readability ratings (\S \ref{sec:relatedwork}). }
\vspace{-.2cm}
\label{tab:summary-datasets}
\end{table*}

Recent advancements in Language Models (LMs) \citep{mt5,xlmr} have enabled the development of neural-based readability assessment methods \citep{rsrs}. Despite the progress made, the absence of a diverse benchmark limits the ability to effectively evaluate how well LM-based methods, whether supervised, unsupervised, or prompting-based, perform across domains and languages. Current evaluation resources for sentence readability assessment suffer from a few crucial shortcomings. First, existing datasets are primarily composed of sentences collected from Wikipedia \cite{tcde,cefr,stajner} or news articles \cite{brunato}. However, LMs have been shown to struggle when handling data from a different domain outside of their training corpus \cite{plank2016non, farahani2021brief, arora-etal-2021-types}. For reliable readability assessment, it's critical for methods to perform well across various textual domains. Hence, a domain-diverse benchmark is essential in assessing model domain generalization. Past work also often utilized document-based readability data as an approximation for sentence-based readability (more in \S \ref{sec:relatedwork}), due to a lack of human readability ratings on individual sentences \cite{rsrs, nprm}. Additionally, there is no existing benchmark for sentence readability assessment that covers a diverse set of language families, limiting the ability to perform cross-lingual evaluation and analysis.

To address these gaps in the field, we introduce {\sc ReadMe++}, a diverse multi-domain dataset for multilingual sentence readability assessment. {\sc ReadMe++} consists of 9757 human-annotated sentences drawn from 112 distinct data sources and covers 5 different languages: Arabic, English, French, Hindi, and Russian (see examples in Figure~\ref{fig:examples-intro}). We focus on readability assessment for second language learners \cite{xia2019text} and thus annotate sentences for their readability level based on the Common European Framework of Reference for Languages (CEFR) scale (\S~\ref{sec:annotation}). 

Using {\sc ReadMe++}, we benchmark a variety of monolingual and multilingual LMs for multi-domain readability assessment in the supervised, unsupervised, and few-shot prompting settings. The domain and language diversity in {\sc ReadMe++} enable us to analyze more effective few-shot prompting (\S~\ref{sec:supervised-methods}) and identify shortcomings in existing unsupervised readability prediction methods, such as the effect of transliterations on their performance in languages with non-Latin script (\S~\ref{sec:unsupervised-methods}). Finally, we show that LMs fine-tuned using {\sc ReadMe++} perform better on unseen domains and exhibit superior cross-lingual transfer capabilities from English to six target languages: Arabic, French, Hindi, Russian, Italian, and German, compared with LMs trained on previous datasets (\S~\ref{sec:analyses}).

\section{Related Work}
\label{sec:relatedwork}

\paragraph{Document-based Readability.} Many datasets used in readability research have only document-level labels, as they were collected from sources (e.g., textbooks) that provide parallel or non-parallel text at varied levels of writing.  These include WeeBit \cite{weebit}, Newsela \cite{newsela}, Cambridge \cite{cambridge}, OneStopEnglish \cite{onestopenglish}, VikiWiki \cite{mrnn}, Slovenian SB \cite{rsrs}, English-Chinese LR \cite{R3}, ALC \cite{ar-corpora}, Gloss \cite{ar-corpora}, ZAEBUC \cite{habash2022zaebuc}, SAMER \cite{alhafni2024samer}, and Philippines Corpus \cite{imperial2023automatic}. While appropriate for assessing document readability, such datasets are suboptimal for sentence-level readability compared to resources with ground-truth readability labels for individual sentences \cite{cripwell2023simplicity}.

\begin{figure}[t]
    \centering
\includegraphics[width=\linewidth]{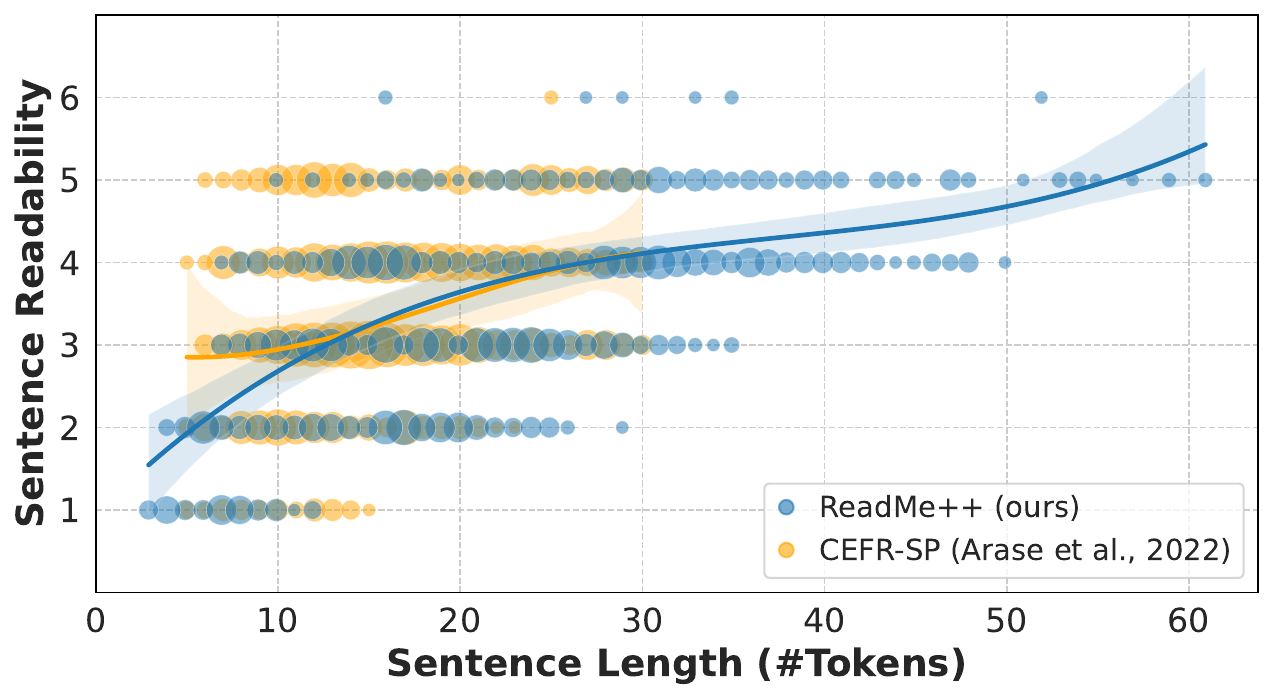}
    \caption{Distribution of sentence lengths across readability levels in the English portion of {\sc ReadMe++}, compared with CEFR-SP \cite{cefr}. {\sc ReadMe++} offers a wider coverage of lengths and readability levels. }
    \label{fig:readme-vs-cefr}
\end{figure}

\begin{table*}[h]
\centering
\setlength{\tabcolsep}{5pt}
\begin{adjustbox}{width=\linewidth}
\begin{tabular}{@{}lclll@{}}
\toprule
 & \multicolumn{1}{l}{} & \multicolumn{3}{c}{\textbf{Examples of Data Sources} --- Full list for all languages in Appendix~\ref{appendix:domains}} \\ \cmidrule(l){3-5} 
\multirow{-2}{*}{\textbf{Domain (Abrv)}} & \multicolumn{1}{l}{\multirow{-2}{*}{\textbf{\#}}} & \multicolumn{1}{c}{\textbf{Arabic (ar)}} & \multicolumn{1}{c}{\textbf{English (en)}} & \multicolumn{1}{c}{\textbf{Hindi (hi)}} \\ \midrule
\rowcolor[HTML]{EFEFEF} 
{\sc Captions} (Cap) & 9 & \textbf{Images} \cite{image-captions-ar} & \textbf{Videos} \cite{video-captions-en} & \textbf{Movies} \cite{open-subtitles} \\
{\sc Dialogue} (Dia) & 7 & \textbf{Open-domain} \cite{open-domain-ar} & \textbf{Negotiation} \cite{negotiation-en} & \textbf{Task-oriented} \cite{task-oriented-hi} \\
\rowcolor[HTML]{EFEFEF} 
{\sc Dictionaries} (Dic) & 2 & \textbf{Dictionaries} (almaany.com) & \textbf{Dictionaries} (dictionary.com) & --- \\
{\sc Entertainment} (Ent) & 4 & \textbf{Jokes} (almrsal.com) & \textbf{Jokes} \cite{jokes-en} & \textbf{Jokes} (123hindijokes.com) \\
\rowcolor[HTML]{EFEFEF} 
{\sc Finance} (Fin) & 3 & --- & \textbf{Finance} \cite{finance} & --- \\
{\sc Forums} (For) & 7 & \textbf{QA Websites} \cite{qa-ar} & \textbf{StackOverflow} \cite{stackoverflow} & \textbf{Reddit} (reddit.com) \\
\rowcolor[HTML]{EFEFEF} 
{\sc Guides} (Gui) & 6 & \textbf{Online Tutorials} (ar.wikihow.com) & \textbf{Code Documentation} (mathworks.com) & \textbf{Cooking Recipes} (narendramodi.in) \\
{\sc Legal} (Leg) & 9 & \textbf{UN Parliament} \cite{un-parallel-corpus} & \textbf{Constitutions} (constitutioncenter.org) & \textbf{Judicial Rulings} \cite{judicial-hi} \\
\rowcolor[HTML]{EFEFEF} 
{\sc Letters} (Let) & 3 & --- & \textbf{Letters} (oflosttime.com) & --- \\
{\sc Literature} (Lit) & 3 & \textbf{Novels} (hindawi.org/books/) & \textbf{History} (gutenberg.org) & \textbf{Biographies} (Public Domain Books) \\
\rowcolor[HTML]{EFEFEF} 
{\sc Medical Text} (Med) & 1 & --- & \textbf{Clinical Reports} \cite{clinical-reports-en} & --- \\
{\sc News Articles} (New) & 2 & \textbf{Sports} \cite{arabic-news} & \textbf{Economy} \cite{english-news} & --- \\
\rowcolor[HTML]{EFEFEF} 
{\sc Poetry} (Poe) & 5 & \textbf{Poetry} (aldiwan.net) & \textbf{Poetry} (poetryfoundation.org) & \textbf{Poetry} (hindionlinejankari.com) \\
{\sc Policies} (Pol) & 7 & \textbf{Olympic Rules} (specialolympics.org) & \textbf{Contracts} (honeybook.com) & \textbf{Code of Conduct} (lonza.com) \\
\rowcolor[HTML]{EFEFEF} 
{\sc Research} (Res) & 15 & \textbf{Politics} (jcopolicy.uobaghdad.edu.iq) & \textbf{Science \& Engineering} (arxiv.org) & \textbf{Economics} (journal.ijarms.org) \\
{\sc Social Media} (Soc) & 3 & \textbf{Twitter} \cite{stanceosaurus} & \textbf{Twitter} \cite{stanceosaurus} & \textbf{Twitter} \cite{stanceosaurus} \\
\rowcolor[HTML]{EFEFEF} 
{\sc Speech} (Spe) & 4 & \textbf{Public Speech} (state.gov/translations) & \textbf{Public Speech} (whitehouse.gov) & \textbf{Ted Talks} (ted.com/talks) \\
{\sc Statements} (Sta) & 6 & \textbf{Quotes} (arabic-quotes.com) & \textbf{Rumours} \cite{stanceosaurus} & \textbf{Quotes} (wahh.in) \\
\rowcolor[HTML]{EFEFEF} 
{\sc Textbooks} (Tex) & 3 & \textbf{Business} (hindawi.org/books/) & \textbf{Agriculture} (open.umn.edu) & \textbf{Psychology} (ncert.nic.in) \\
{\sc User Reviews} (Rev) & 12 & \textbf{Products} \cite{reviews-ar} & \textbf{Books} (goodreads.com) & \textbf{Movies} (hindi.webdunia.com) \\

\rowcolor[HTML]{EFEFEF} 
{\sc Wikipedia} (Wik) & 1 & \textbf{Wikipedia} (wikipedia.com) & \textbf{Wikipedia} (wikipedia.com) & \textbf{Wikipedia} (wikipedia.com) \\ \midrule
\multicolumn{1}{r}{\textbf{Total}} & 112 &  &  &  \\ \bottomrule
\end{tabular}
\end{adjustbox}
\caption{List of domains and example data sources in {\sc ReadMe++} (see full list for all 5 languages in Appendix \ref{appendix:domains}).}
\vspace{-.1cm}
\label{tab:dataset-domains}
\end{table*}

\paragraph{Sentence-based Readability.} Only a few existing datasets \cite{mtde,stajner,brunato,tcde} were created by manually annotating individual sentences for their level of readability (see Table~\ref{tab:summary-datasets}). However, these sentence-level annotated datasets are largely limited to high-resource English and European languages that use the Latin script. They are also collected from one or a few data sources and are thus insufficient for studying the robustness of readability assessment methods across text domains. Further, these past datasets are annotated with various rating scales that do no have a clear readability grounding. The recent CEFR-SP dataset \cite{cefr} adopts the 6-level CEFR scale for annotation, which grounds sentence readability in the language capability of a second language learner. However, CEFR-SP only contains English sentences from Wikipedia, Newsela \cite[leveled news articles]{newsela}, and SCoRE \cite[textbooks for learning English]{score}. In comparison, our work highlights the importance of both domain and language coverage, resulting in more data diversity (see Figure~\ref{fig:readme-vs-cefr}). {\sc ReadMe++} covers 112 different data sources and is annotated at the sentence level in 5 languages.

\paragraph{Multilingual Readability Assessment.} 
Several works have leveraged neural approaches for multilingual readability assessment. Many adopt fine-tuning strategies of transformer LMs \cite{mrnn, R10, R6, R8, R4, R2}. However, training data is often unavailable except in a few high-resource languages. Other works explored cross-lingual transfer strategies \cite{imperial2023automatic}, demonstrating effective transfer from English to French/Spanish \cite{nprm} and Chinese \cite{R3}. The work of \citet{rsrs} proposed an unsupervised approach that leverages an LM's distribution to compute a likelihood-based sentence readability score. The majority of these past studies have used document-based readability datasets. Using our dataset, we benchmark various LMs in the supervised, unsupervised, and few-shot prompting settings in diverse language scripts (i.e., Arabic, Latin, Brahmic, and Cyrillic). We show that LMs trained using the English portion of {\sc ReadMe++} perform better cross-lingual transfer to 6 target languages compared to models trained on previous datasets.

\section{Constructing {\sc ReadMe++} Corpus}

We present the detailed procedure for constructing the {\sc ReadMe++} corpus. To maximize the diversity of domains, we identified 112 data sources that are either with open licenses or shareable for non-commercial purposes (see Table \ref{tab:dataset-domains}). A total of 9757 sentences (1945 Arabic, 1669 French, 2861 English, 1524 Hindi, 1758 Russian) were sampled from these sources and then manually annotated. {\sc ReadMe++} supports multilingual, cross-lingual, and cross-domain experiments (\S \ref{sec:benchmarking}).

\subsection{Data Collection}
\label{sec:dataset}

\paragraph{Selecting Diverse Data Sources.} Our data collection process varies per source and can be categorized into four approaches: \textbf{(1)} obtaining content directly from a website (e.g., Wikipedia), \textbf{(2)} extracting text from sources in PDF format (e.g., contract templates, reports, etc.), \textbf{(3)} sampling text from existing datasets (e.g., dialogue, user reviews, etc.), or \textbf{(4)} manually collecting sentences (e.g., dictionary examples, etc.). Collection details per domain are provided in Appendix~\ref{appendix:domains}. For each domain, we collected the available texts from one or more data sources and then sampled 50 paragraphs per domain. We increased the sampling rate to 100 for unstructured sources such as PDFs since they are likely to return text not useful for annotation (e.g., headers, titles, references, etc.) that needs to be filtered out. From each paragraph, we sample one sentence that we use for readability annotation. Lastly, we perform manual quality checking to filter out any low-quality sentences and sentences that contain toxic, hateful, or offensive language.

\paragraph{Considering the Influence of Contexts.} In addition to the sampled sentences, we collect up to three preceding sentences as context if available. Many of the sampled sentences could be placed in the body of a paragraph. We provided annotators with optional access to context in case they needed to know the context in which a sentence appears. Such cases have not been adequately considered in previous work; for example, \citet{cefr} collected only the first sentence in a paragraph. We provide additional results in Appendix~\ref{app:context-effect} where context was provided to LMs during fine-tuning.

\subsection{Readability Annotation}
\label{sec:annotation}

\paragraph{Using the CEFR Standards.}  
Previous works on sentence-level readability have used various rating scales such as 0-100 \cite{mtde}, 3-point \cite{stajner}, or 7-point \cite{tcde,brunato} scales. However, these scales are prone to annotator subjectivity due to the lack of a clear readability grounding. Instead, following \citet{cefr}, we adopt the Common European Framework of Reference for Languages (CEFR), which defines the language ability of a person on a 6-point scale (1$_{(A1)}$, 2$_{(A2)}$, 3$_{(B1)}$, 4$_{(B2)}$, 5$_{(C1)}$, 6$_{(C2)}$), where A is for basic, B for independent, and C for proficient. Each level of the scale is grounded by can-do descriptors of a language learner, which act as a guide for annotators (see CEFR level descriptors in Appendix \ref{app:cefr-levels}).

\paragraph{Rank-and-Rate Annotation.} Rating each sentence independently on a scale of readability comes with the drawback of annotators eventually not differentiating between different sentences. This results in most samples being labeled within one or two levels, limiting their usefulness for statistical analyses \cite{rank-rate}. Instead of rating alone as in prior works, we utilize a Rank-and-Rate approach \cite{maddela-etal-2023-lens} for readability annotation, which mitigates independent sentence rating issues by providing comparative texts. We randomly group sentences into batches of 5 and ask annotators to first rank sentences of a batch from most to least readable and then rate each sentence individually on the 6-point CEFR scale. By comparing and contrasting sentences within a batch, annotators can better differentiate between the readability of different sentences and produce less subjective ratings. In our initial pilot studies, we found that annotators express a better experience when using the rank-and-rate framework and achieve higher agreements compared with rating alone. Our interface is shown in Appendix~\ref{appendix:interface}.

\begin{table}[t]
\centering
\small
\begin{tabular}{@{}clcc@{}}
\toprule
\multicolumn{2}{c}{\textbf{Dataset}} & $\alpha$ & $\rho$ \\ \midrule
\multirow{5}{*}{\textbf{{\sc ReadMe++}}} & \textbf{Arabic} & 0.67 & 0.78 \\
 & \textbf{English} & 0.78 & 0.81 \\
 & \textbf{French} & 0.76 & 0.78 \\
 & \textbf{Hindi} & 0.67 & 0.71 \\
 & \textbf{Russian} & 0.68 & 0.72 \\ \midrule
\multirow{2}{*}{\begin{tabular}[c]{@{}c@{}} CEFR-SP\\ \small{\cite{cefr}}\end{tabular}} & \textbf{WikiAuto} & 0.66 & 0.73 \\
 & \textbf{SCoRe} & 0.44 & 0.66 \\ \bottomrule
\end{tabular}
\caption{Annotator agreements measured by Krippendorff's alpha ($\alpha$) and Pearson Correlation ($\rho$). The agreements reached in CEFR-SP \cite{cefr} are provided for comparison.}
\label{tab:annotator-agreement}
\end{table}


\paragraph{Annotator Selection.} We take several steps to ensure the quality of our annotations. First, four of our authors who can speak each language provided the first set of annotations. We then hired two additional annotators for each language, who were university students who can speak the language and had linguistic annotation experience, or annotators we hired through Prolific. Annotators were paid at rates of \$16-18/hour. When recruiting annotators, we first conducted training sessions to familiarize them with the CEFR scale and the annotation framework. We then gave each candidate a batch of 250 sentences and only proceeded with candidates who achieved a sufficient enough correlation (> 0.7) with the first set of annotations.

\paragraph{Inter-annotator Agreement.} We report the Krippendorff's alpha ($\alpha$) and average Pearson Correlation ($\rho$) between the three annotators for each language in Table~\ref{tab:annotator-agreement}. High agreements are achieved by our annotators \cite{artstein2008inter}, on par with the past work of \citet{cefr}. We perform majority voting on the three annotations to obtain a final rating that we use in our experiments.

\section{Benchmarking Experiments}
\label{sec:benchmarking}
\begin{figure}[t]
    \centering
\includegraphics[width=\linewidth]{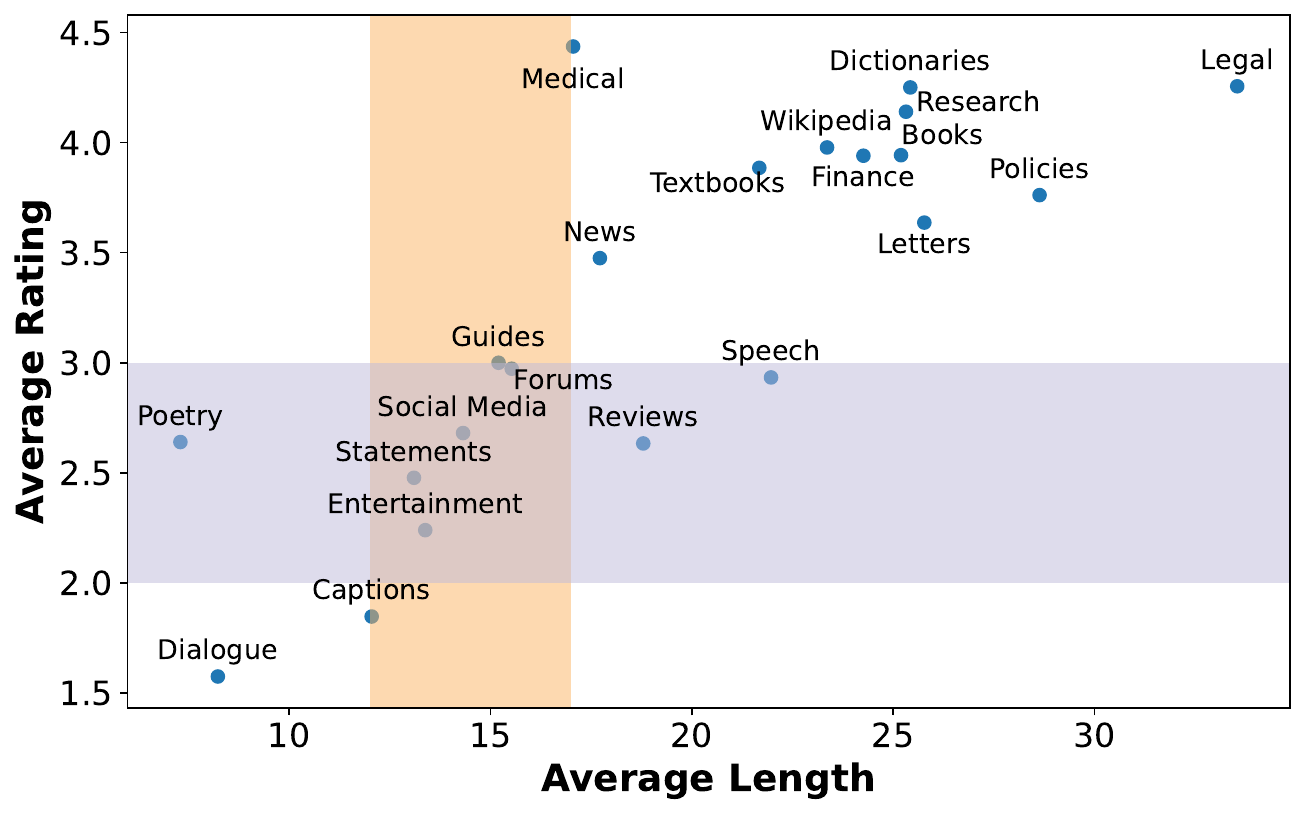}

    \caption{Average readability rating and sentence length per domain in the English portion of {\sc ReadMe++}. Domain diversity presents additional challenges for readability assessment. Certain domains may be within the same readability range (e.g. \textcolor{lightviolet}{[2, 3] that corresponds to A2 and B1 levels}) but have varying lengths, while sentences within a length range (e.g. \textcolor{lightorange}{[12, 17] tokens}) could be spread across the whole readability spectrum.}
    \label{fig:readme-domains}
\end{figure}

As shown in Figures \ref{fig:readme-vs-cefr} and \ref{fig:readme-domains}, the {\sc ReadMe++} corpus offers a diverse coverage of domains, readability levels, and sentence lengths, making it an ideal testbed for evaluating readability assessment methods. We benchmark supervised, unsupervised, and few-shot approaches using recently developed LMs. We use the same random train/valid/test split (detailed statistics in Appendix~\ref{app:split-stats}) based on a 60/10/30\% ratio per domain for all experiments, except the domain generalization study in \S \ref{sec:analyses}.


\subsection{Supervised \& Prompting Methods}
\label{sec:supervised-methods}

\paragraph{Supervised.} We fine-tune LMs to classify sentence readability. We compare multilingual models, \textbf{mBERT} \cite{bert} and \textbf{XLM-R} \cite{xlmr}, to monolingual models that include \textbf{BERT} \cite{bert} for English, \textbf{AraBERT} \cite{arabert} and \textbf{ArBERT} \cite{arbert} for Arabic, \textbf{CamemBERT} for French \cite{martin2020camembert}, and \textbf{RuBERT} \cite{rubert} for Russian. For Hindi, we use \textbf{MuRIL} \cite{muril} and \textbf{IndicBERTv2} \cite{kakwani2020indicnlpsuite}, both pre-trained on 12 Indian languages. We also consider encoder-decoder LMs, \textbf{mT5} \cite{mt5}, \textbf{Aya101} \cite{ustun2024aya}, and \textbf{AraT5} \cite{arat5}.  We fine-tune for 20 epochs using the cross-entropy loss and the Adam optimizer and tune the learning rate in the set $\{1e^{-5}, 1e^{-6}, 1e^{-7}\}$. We select checkpoints based on the best performance on the validation set. We report the average of 5 runs with different random initialization seeds.

\paragraph{Prompting.} We perform in-context learning using \textbf{GPT3.5}, \textbf{GPT4} (Apr 2024), \textbf{Llama2-7b} \cite{llama2}, \textbf{Llama3.1-8b} \cite{dubey2024llama}, and \textbf{Aya23-8b} \cite{aryabumi2024aya}. We provide LMs with a definition of readability and the descriptors of the six CEFR levels. We show the model five randomly sampled in-context examples from the train set and their corresponding CEFR levels, then ask the model to assess the readability of a new sentence based on the CEFR scale. Prompt details can be found in Appendix~\ref{app:prompting}.

\begin{figure*}[t]
    \centering
    \includegraphics[width=\linewidth]{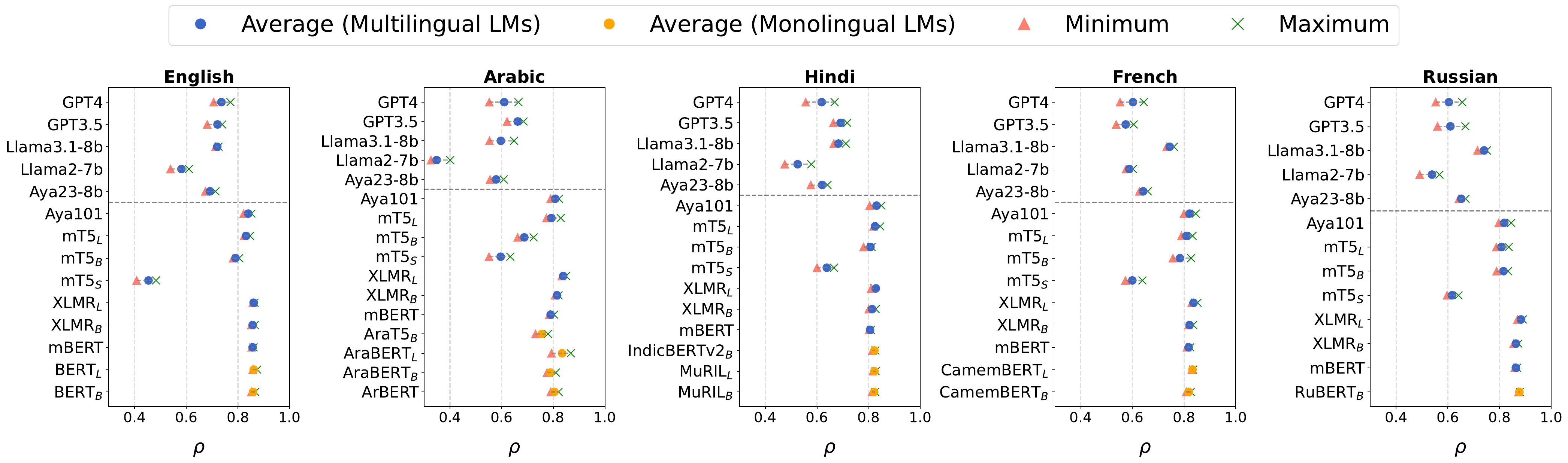}
    \vspace{-.5cm}
    \caption{Pearson correlation ($\rho$) of \textbf{fine-tuned} multilingual and monolingual LMs, as well as \textbf{prompted} GPT3.5, GPT4, Aya23-8b, Llama2-7b, and Llama3.1-8b models with 5-shot examples, on the test set of {\sc ReadMe++}. The small ($_S$), base ($_B$), and large ($_L$) sizes of the models are used. We report the min/max/average of performance across 5 runs using random seeds for fine-tuning initialization, or random sets of demonstrations in prompting.}
    \label{fig:sup-results}
\end{figure*}

\subsubsection{Results} The results are shown per language in Figure~\ref{fig:sup-results}, where we report the Pearson Correlation ($\rho$) between the predictions and the ground-truth labels. Additional metrics are reported in Appendix~\ref{app:main-f1-scores}. 

\paragraph{A gap exists between fine-tuning and few-shot performance.}  Fine-tuned models were able to achieve high correlation levels in the 0.7-0.9 range, with larger models showing improved performance. Overall, mT5$_{L}$ was among the best-performing fine-tuned models across all languages. However, the performance of prompted causal models with 5-shot examples was lower than that of fine-tuned models in all languages.

\begin{figure}[t]
    \centering
    \includegraphics[width=\linewidth]{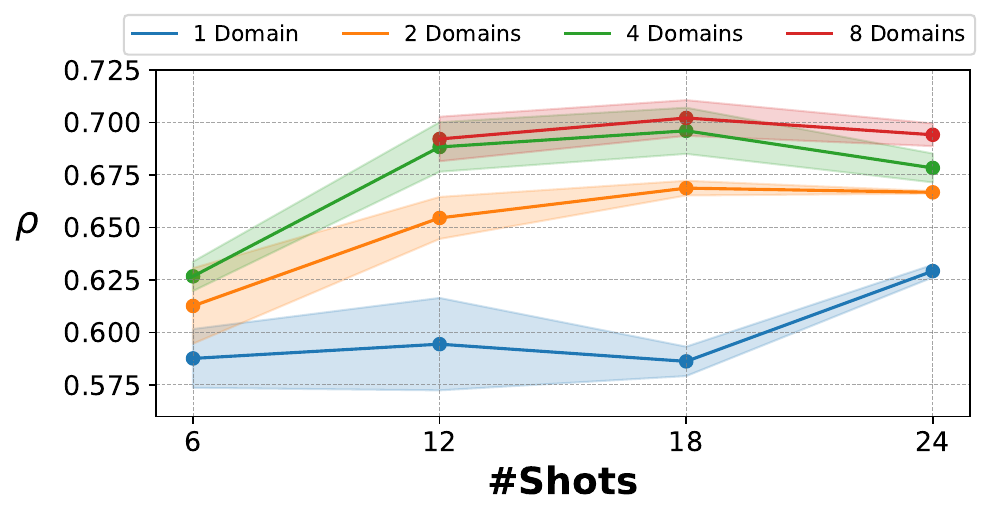}
    \caption{Effect of domain diversity of in-context examples on Llama2-7b performance on {\sc ReadMe++} (\textit{en}). Correlation is greatly improved when examples are sampled from an increasing number of domains.}
    \label{fig:prompt-analysis}
\end{figure}

\paragraph{Domain diversity of in-context examples improves few-shot performance.} We analyze the effect of the domain diversity of the few-shot examples on prompting performance. We prompt Llama2 by sampling examples from 1, 2, 4, and 8 domains. The domains from which the examples are sampled are also randomly sampled for each test sentence. The average correlation from 5 runs is shown in Figure~\ref{fig:prompt-analysis}, for an increasing number of shots. The performance gain from increasing domain diversity is clearly observed, with correlation improving all cases, reaching slightly above 0.7 in the best case. This improvement also outweighs the gains from increasing the number of shots, highlighting the importance of domain diversity.

\begin{figure*}[t]
    \centering
    \includegraphics[width=\linewidth]{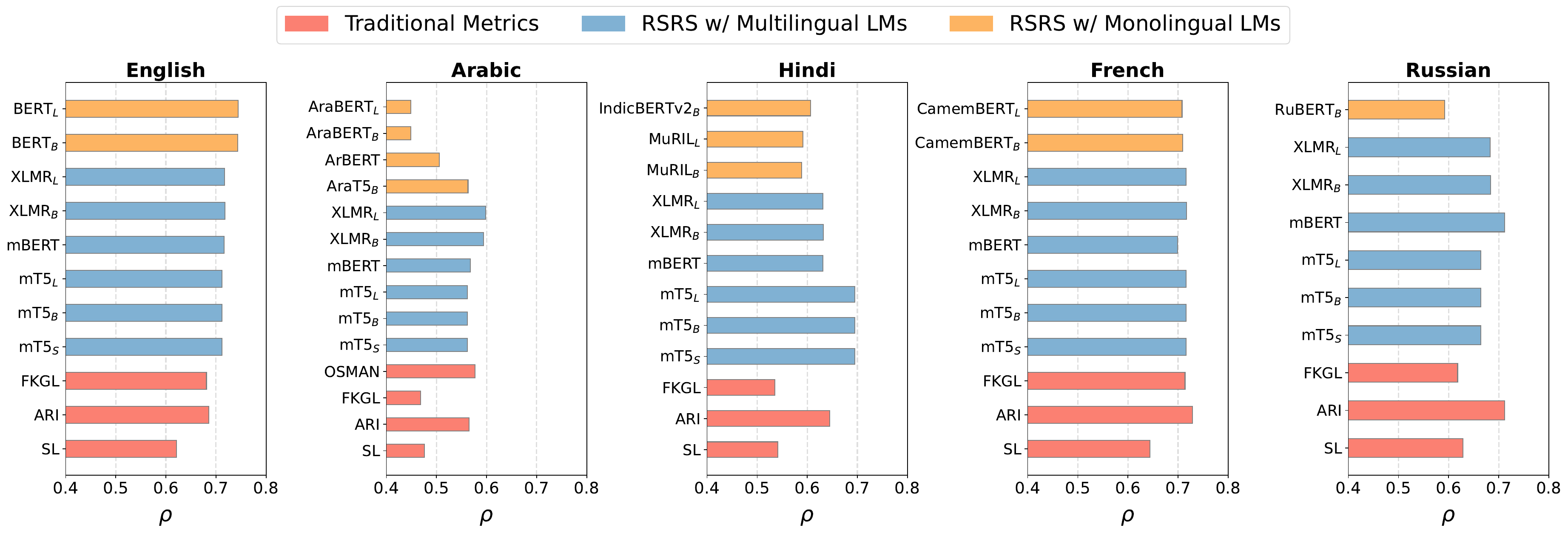}
    \vspace{-.4cm}
    \caption{Pearson correlation ($\rho$) of \textbf{unsupervised} readability measurements on the test set of {\sc ReadMe++}, including RSRS \cite{rsrs} which leverages conditional word probabilities estimated by LMs. RSRS which uses multilingual LLMs performs better than RSRS which uses monolingual models in languages with non-Latin scripts.}
    \vspace{-.2cm}
     \label{fig:unsup-results}
\end{figure*}

\begin{figure}[t]
    \centering
    \includegraphics[width=\linewidth]{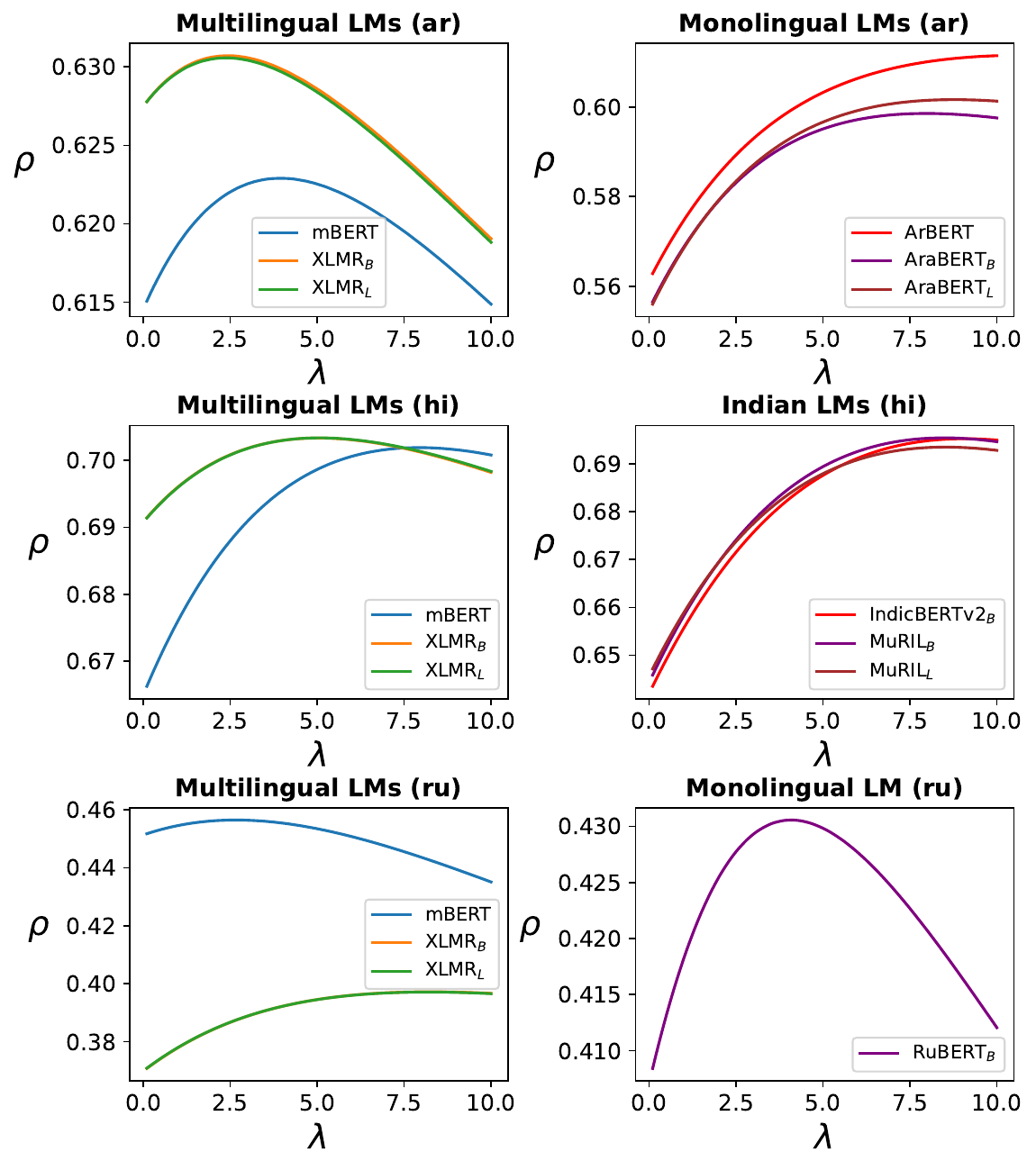}
    \caption{Effect of increasing the penalty factor ($\lambda$) on the Pearson correlation ($\rho$) between RSRS scores and human ratings for Arabic, Hindi and Russian sentences that contains transliterations. The plot shows a clear improvement in correlation as $\lambda$ increases, which is more significant for monolingual than multilingual models.}
    \label{fig:effect-transliteration}
\end{figure}

\subsection{Unsupervised Methods}
\label{sec:unsupervised-methods}

In the unsupervised setting, we leverage the LM distribution to compute a readability score without training. We also compare with several traditional length-based readability formulas. 

\paragraph{LM-based Metrics.} We use the Ranked Sentence Readability Score (\textbf{RSRS}) proposed by \citet{rsrs} which combines LM statistics with the sentence length. It computes a weighted sum of the individual word losses as follows:
\begin{equation}
    \mathrm{RSRS} = \frac{\sum_{i=1}^{S}{[\sqrt{i}]^{\alpha}.\mathrm{WNLL}(i)}}{S},
\end{equation}

\noindent where $S$ is the sentence length, $i$ is the rank of the word after sorting each Word's Negative Log Loss (WNLL) in ascending order. Words with higher losses are assigned higher weights, increasing the total score and reflecting less readability. $\alpha$ is equal to 2 when a word is an Out-Of-Vocabulary (OOV) token and 1 otherwise, assuming that OOV tokens represent rare, difficult words and thus are assigned higher weights by eliminating the square root. The WNLL is computed as follows:
\begin{equation}
    \mathrm{WNLL} = -(y_t \log y_p + (1-y_t)\log(1-y_p)),
\end{equation}

\noindent where $y_p$ is the predicted distribution by the LM, and $y_t$ is the true distribution where the word appearing in the sequence holds a value of 1 while all other words have a value of 0.

\paragraph{Traditional Readability Metrics.} We compare to several common traditional readability metrics \cite{ehara2021evaluation}, which are based on word and sentence lengths. Specifically, we use the Sentence Length \textbf{(SL)}, Automated Readability Index \textbf{(ARI)} \cite{ari}, Flesch-Kincaid Grade Level \textbf{(FKGL)} \cite{fkgl}, and Open Source Metric for Measuring Arabic Narratives \textbf{(OSMAN)} \citep{osman}. The formulas for these metrics are provided in Appendix~\ref{appendix:feature-based-metrics}.

\subsubsection{Results}

The results achieved by unsupervised methods are shown in Figure~\ref{fig:unsup-results}. We find that \textbf{LM-based RSRS scores achieve better correlation than traditional readability metrics in English. This was not the case for other languages, where performance was model-dependent.} Interestingly, for languages with non-Latin script (Arabic, Hindi, Russian), we find that RSRS scores computed via monolingual LMs achieve noticeably lower correlations compared to multilingual LMs. The RSRS metric (\S \ref{sec:unsupervised-methods} Eq. 1) assumes that all unseen words by the LM's tokenizer are rare, difficult words that should be assigned higher weights. However, these could also be transliterations from other languages (e.g., names of new politicians or artists, emerging diseases, historical figures, etc.) that the LM never saw during pre-training. We hypothesize that this design choice in RSRS degrades its performance on languages with non-Latin script since many of these transliterated words do not add to the difficulty level of the sentence for humans.

\paragraph{Unsupervised LM-based RSRS struggle with transliterations.}  To test the impact of transliterated words on RSRS scores, we asked Arabic, Hindi, and Russian annotators to indicate if a sentence contains transliterated words when annotating. This resulted in 320 sentences with transliterations in Arabic (16.45\% of Arabic data), 561 sentences in Hindi (36.81\% of Hindi data), and 120 sentences in Russian (6.82\% of Russian data). We penalized the RSRS scores of those sentences by a factor $\frac{\lambda}{S}$, where $\lambda$ is a penalty factor and $S$ is the length of the sentence. We compute the correlation with human labels for an increasing penalty $\lambda$ to analyze whether decreasing those scores results in a higher correlation since we assume transliterations cause RSRS scores to
be unreasonably high. The results are shown in Figure~\ref{fig:effect-transliteration} for 0.1 increments of $\lambda$. The trends corroborate with our hypothesis, where correlation increases as the penalty becomes higher up to a certain level. The improvement reaches up to 6-7\% for monolingual LMs. Multilingual LMs (improvements of 1-3\%) were less affected, indicating their greater robustness to transliterations. This underscores the need for careful consideration of transliterations in future research.

\begin{table*}[]
\centering
\small
\begin{adjustbox}{width=0.87\linewidth}
\renewcommand{\arraystretch}{0.87}
\begin{tabular}{@{}clcrrrrr@{}}
\toprule
\multicolumn{1}{l}{} & \multicolumn{1}{c}{\multirow{2}{*}{\textbf{\#Unseen Domains (\#Data Sources)}}} & \multirow{2}{*}{\textbf{\#train/val}} & \multirow{2}{*}{\textbf{\#test}} & \multicolumn{2}{c}{\textbf{ReadMe++}} & \multicolumn{2}{c}{\textbf{CEFR-SP}} \\ \cmidrule(l){5-8} 
\multicolumn{1}{l}{} & \multicolumn{1}{c}{} &  &  & \multicolumn{1}{c}{\textbf{F1}} & \multicolumn{1}{c}{\textbf{$\rho$}} & \multicolumn{1}{c}{\textbf{F1}} & \multicolumn{1}{c}{\textbf{$\rho$}} \\ \midrule
\multirow{4}{*}{\textbf{English}} & 2 (7): Wik, Res & 1995 / 235 & 631 & \textbf{37.57} & \textbf{0.611} & 20.95 & 0.439 \\
 & 4 (7): Let, Ent, Soc, Gui & 2285 / 267 & 309 & \textbf{40.16} & \textbf{0.761} & 24.91 & 0.649 \\
 & 6 (14): Res, Fin, Sta, Ent, Dia, New & 1885 / 221 & 755 & \textbf{34.61} & \textbf{0.780} & 20.69 & 0.517 \\
 & 8 (25): Pol, Cap, Sta, Res, Rev, Leg, Soc, Poe & 1653 / 191 & 1017 & \textbf{43.88} & \textbf{0.828} & 23.80 & 0.690 \\ \midrule \midrule
\multicolumn{1}{l}{} & \multicolumn{1}{c}{\multirow{2}{*}{\textbf{\#Unseen Domains (\#Data Sources)}}} & \multirow{2}{*}{\textbf{\#train/val}} & \multirow{2}{*}{\textbf{\#test}} & \multicolumn{2}{c}{\textbf{ReadMe++}} & \multicolumn{2}{l}{\textbf{ALC Corpus}} \\ \cmidrule(l){5-8} 
\multicolumn{1}{l}{} & \multicolumn{1}{c}{} &  &  & \textbf{F1} & \multicolumn{1}{c}{\textbf{$\rho$}} & \textbf{F1} & \multicolumn{1}{c}{\textbf{$\rho$}} \\ \midrule
\multirow{4}{*}{\textbf{Arabic}} & 2 (2): Tex, New & 1540 / 180 & 225 & \textbf{47.54} & \textbf{0.626} & 6.80 & -0.208 \\
 & 4 (7): Poe, Gui, Ent, Dia & 1457 / 173 & 315 & \textbf{39.24} & \textbf{0.683} & 7.27 & -0.043 \\
 & 6 (11): For, New, Spe, Cap, Wik, Res & 910 / 106 & 929 & \textbf{34.47} & \textbf{0.609} & 10.25 & 0.083 \\
 & 8 (13): Ent, For, Leg, Spe, Wik, Dia, Poe, Res & 918 / 109 & 918 & \textbf{29.56} & \textbf{0.523} & 6.79 & 0.144 \\ \bottomrule
\end{tabular}
\end{adjustbox}
\caption{Supervised mBERT-based readability model fine-tuned on our {\sc ReadMe++} corpus achieve much better performance on unseen domains than the same model trained on existing datasets, namely CEFR-SP \cite{cefr} for English and the ALC Corpus \cite{ar-corpora} for Arabic.}
\label{tab:unseen-domains}
\end{table*}

\begin{table}[]
\centering
\begin{adjustbox}{width=\linewidth}
\renewcommand{\arraystretch}{0.9}
\begin{tabular}{llrrrrrr@{}}
\toprule
\multicolumn{1}{l}{\multirow{2}{*}{\textbf{\textbf{src} $\rightarrow$ \textbf{tgt}}}} & \multicolumn{2}{c}{\textbf{ReadMe++}} & \multicolumn{2}{c}{\textbf{CEFR-SP}} & \multicolumn{2}{c}{\textbf{CompDS}} \\ \cmidrule(l){2-7} 
\multicolumn{1}{c}{} & \multicolumn{1}{c}{\textbf{F1}} & \multicolumn{1}{c}{\textbf{$\rho$}} & \multicolumn{1}{c}{\textbf{F1}} & \multicolumn{1}{c}{\textbf{$\rho$}} & \multicolumn{1}{c}{\textbf{F1}} & \multicolumn{1}{c}{\textbf{$\rho$}} \\ \midrule
\textbf{en} $\rightarrow$ \textbf{ar} & \textbf{31.48} & \textbf{0.606} & 8.81 & 0.071 & 5.99 & 0.322 \\ 
\textbf{en} $\rightarrow$ \textbf{hi} & \textbf{23.87} & \textbf{0.702} & 13.15 & 0.267 & 10.38 & 0.381 \\ 
\textbf{en} $\rightarrow$ \textbf{fr} & \textbf{30.29}  & \textbf{0.768}  &  11.06 & -0.026  & 5.92 & 0.335 \\ 
\textbf{en} $\rightarrow$ \textbf{ru} & \textbf{24.60}  & \textbf{0.760}  & 15.69  &  0.173 & 10.33  & 0.412 \\ 
\textbf{en} $\rightarrow$ \textbf{it} & \textbf{14.68} & \textbf{0.239} & 9.88 & -0.043 & 10.06 & 0.099 \\ 
\textbf{en} $\rightarrow$ \textbf{de} & \textbf{22.19} & \textbf{0.701} & 10.00 & -0.092 & 11.84 & 0.408 \\ \bottomrule
\end{tabular}
\end{adjustbox}
\caption{Zero-shot cross-lingual transfer results using XLMR$_{L}$. LMs fine-tuned on English data (en) of {\sc ReadMe++} significantly outperform LMs fine-tuned with CEFR-SP \cite{cefr} or CompDS \cite{brunato} in transfer to Arabic (ar),  Hindi (hi), French (fr), Russian (ru), Italian (it), and German (de).}
\label{tab:cross-lingual}
\end{table}

\section{Cross-Domain Cross-Lingual Analyses}
\label{sec:analyses}

We test the ability of LMs trained on {\sc ReadMe++} to generalize to unseen domains (\ref{subsec:unseen-domains}) and transfer to other languages (\ref{subsec:cross-ling-transfer}) compared with models trained on previous datasets.

\subsection{Performance on Unseen Domains}
\label{subsec:unseen-domains}

To test how well fine-tuned models perform on unseen domains, we create new train/val/test splits from {\sc ReadMe++} by removing an increasing number of randomly sampled domains from the dataset (Table~\ref{tab:unseen-domains}). We use the sentences from the removed domains as the test set and use the rest of the dataset for training and validation. For direct comparison, we randomly sample the same amount of train/val sentences in each experiment from the open-sourced Wikipedia-based portion of CEFR-SP \cite{cefr} to fine-tune mBERT models. We evaluate on the unseen domains test set from {\sc ReadMe++}. The results in Table~\ref{tab:unseen-domains} show that \textbf{models fine-tuned using {\sc ReadMe++} achieve good performance on unseen domains and outperform models trained using CEFR-SP}, demonstrating the advantage of domain diversity in {\sc ReadMe++}.

We perform the same experiments in Arabic by comparing to the ALC Corpus \cite{ar-corpora}, which is labeled on 5-scale CEFR levels (A1, A2, B1, B2, C). We convert the labels in {\sc ReadMe++} to the same scale of ALC Corpus by combining C1 and C2 into C and then perform a 5-way classification. We observe the same trend, where models trained using the Arabic portion of {\sc ReadMe++} achieve good performance on unseen domains and outperform models trained on ALC.

\subsection{Performance on Cross-lingual Transfer}
\label{subsec:cross-ling-transfer}


We perform zero-shot cross-lingual transfer from English to 6 different languages by fine-tuning multilingual models using the English subset of {\sc ReadMe++}. For comparison, we also fine-tune on the same number of train/valid sentences that we randomly sample from the open-sourced Wikipedia-based portion of CEFR-SP \cite{cefr} and the full English CompDS \cite{brunato} corpora. We evaluate on the Arabic, Hindi, French, and Russian test sets from {\sc ReadMe++}, as well as Italian CompDS \cite{brunato} and German TextComplexityDE \cite{tcde}.  Since CompDS and TextComplexityDE rate on scales from 1-7 instead of 1-6 but have only a few level-7 sentences, we merged their level 6 and 7 together. The results are shown in Table~\ref{tab:cross-lingual} for XLMR$_{L}$, where we find that \textbf{the model fine-tuned using {\sc ReadMe++} performs much better cross-lingual transfer across all tested languages} compared to models fine-tuned using CEFR-SP or CompDS, reaching high correlation values of 0.7 in most languages. In several cases, training on {\sc ReadMe++} leads to a 50\% increase in performance. This trend is also observed across several models which we report in Appendix~\ref{app:cross-lingual-transfer}.

\section{Conclusion}

We introduced {\sc ReadMe++}, a multi-domain dataset for multilingual sentence readability assessment. {\sc ReadMe++} provides 9757 sentences in Arabic, English, French, Hindi, and Russian that are collected from 112 different data sources and annotated by humans based on the CEFR scale. We showed that LMs trained using {\sc ReadMe++} achieve strong performance across different textual domains and perform well in cross-lingual transfer from English to 6 target languages, outperforming models trained on previous datasets. By releasing {\sc ReadMe++}, we hope to encourage and enable the development and evaluation of more effective and robust methods for multilingual sentence readability assessment.

\section*{Limitations}

{\sc ReadMe++} offers a diversity of text domains in multiple languages. Most domains in our dataset include texts in all the languages we considered, with a few exceptions where openly accessible data was not available in every language. The medical text domain, which consists of clinical reports, is only available in English. However, medical-related texts in other languages are covered within other domains, such as Research and Wikipedia.

In our experiments on cross-lingual transfer, we showed that models fine-tuned on {\sc ReadMe++} transfer well to other languages and outperform models trained on previous datasets. However, our dataset does not cover low-resource languages, which limits the ability to perform evaluation in such scenarios. Future work can extend {\sc ReadMe++} to include such languages. We will be releasing our rank-and-rate annotation interface that will enable easy extensions of our resource to additional languages by the research community.

We analyzed how transliterations can negatively impact the performance of the LM-based RSRS unsupervised metric due to its approach to handling rare words. However, certain rare words such as jargon and complex terminology could well add to the difficulty of a sentence. The language and domain diversity of our resource will encourage future studies to make a more in-depth exploration of this particular point and enable the development and evaluation of better unsupervised metrics.

\section*{Ethical Considerations}
We are committed to upholding ethical standards in constructing and disseminating the {\sc ReadMe++} corpus. To ensure the integrity of our data collection process, we have made our best effort to obtain data from sources that are available in the public domain, released under Creative Commons or similar licenses, or can be used freely for personal and non-commercial purposes according to the resource's Terms and Conditions of Use. These sources include public domain books, publicly available documents/reports, and publicly available datasets. We use a small number of randomly sampled sentences for academic research purposes, specifically for labeling sentence readability. We have included a full list of licenses and terms of use for each source in Appendix~\ref{appendix:license}. We would like to note that two of the sources we used require access permission from the original authors, specifically the i2b2/VA \cite{clinical-reports-en} and Hindi Product Reviews \cite{review-product-hi} datasets. Therefore, sentences and annotations from these sources will not be shared with the community unless access permission has been obtained from the original authors.

Every annotator was informed that their annotations were being used to create a dataset for readability assessment. When collecting sentences from social media and forums, we have excluded any sampled sentences containing offensive/hateful speech, stereotypes, or private user information.

\section*{Acknowledgments}
The authors would like to thank Nour Allah El Senary, Govind Ramesh, Suraj Mehrotra, and Ryan Punamiya for their help in data annotation. This research is supported in part by the NSF awards IIS-2144493 and IIS-2112633, NIH award R01LM014600, ODNI and IARPA via the HIATUS program (contract 2022-22072200004). The views and conclusions contained herein are those of the authors and should not be interpreted as necessarily representing the official policies, either expressed or implied, of NSF, NIH, ODNI, IARPA, or the U.S. Government. The U.S. Government is authorized to reproduce and distribute reprints for governmental purposes notwithstanding any copyright annotation therein.

\bibliography{references}

\newpage

\newpage

\clearpage

\appendix

\section{More details about {\sc ReadMe++}}
\label{appendix:domains}

\subsection{Textual Domains}

This section provides a description of how sentences were collected from each domain of {\sc ReadMe++}. Table~\ref{tab:dataset-statistics} shows statistics of the corpus and Table~\ref{tab:dataset-sources} summarizes the sources from which data was collected for each domain in each language, including publicly available web resources or open-source datasets.

\begin{itemize}
    \item {\sc Wikipedia}: Wikipedia is an attractive source of multilingual text since most articles are available in a large number of languages. Further, articles belong to a variety of topics where writing style and technicality differ significantly. We select 9 Wikipedia topics and, from each, randomly sample 5 different articles that discuss a certain sub-topic within that topic. For example, an article on \textit{“Information Theory”} belongs to the \textit{“Technology”} topic. We scrape the Arabic, English, French Hindi, and Russian versions of each article.
    
    \item {\sc News Articles}: We leverage resources used for news category classification research, which we find publicly available datasets for in Arabic \cite{arabic-news} and English \cite{english-news}. No similar public resource was found for the other languages.

    \item {\sc Research}:  We collect text from medical, law, politics, and economics research papers in each language if available. We use open-access research archives such as arxiv\footnote{arxiv.org} or HAL\footnote{hal.science}. We also search for open-access research articles published under a Creative Commons license on Google Scholar using the same keyword in each language. We notice that research papers from natural sciences or technology are much less frequent in non-English languages as most researchers in those areas publish their work in English.

    \item {\sc Literature}: We collect sentences from different types of literature \textit{(Novels, History, Biographies, Children's Stories)} using books that are in the public domain. For English, French, and Russian, we use Project Gutenberg\footnote{gutenberg.org} that archives old books for which U.S. copyright has expired. For Arabic, we use Hindawi Books\footnote{hindawi.org} which provide free Arabic books in many genres and topics. For Hindi, the law in India states that the copyright terms of books end 60 years after the death of an author and comes under the public domain\footnote{https://copyright.gov.in/Documents/handbook.html}. Similar laws for most countries of the world are present with varying number of years\footnote{en.wikipedia.org/wiki/List\_of\_countries\%27\_copyright\_lengths}. We thus manually search for books in Hindi whose copyrights have expired according to these lengths. For example, we used Hindi novels by Premchand, Sarat Chandra Chattopadhyay, Rabindranath Tagore and Devaki Nandan Khatri.
    
    \item {\sc Textbooks}: Textbooks are obtained from the Open Textbook Library\footnote{open.umn.edu/opentextbooks/books} for English and Hindawi Books for Arabic which provide openly licensed textbooks. For Hindi textbooks, we use publicly available school textbooks from the National Council of Educational Research and Training in India \footnote{ncert.nic.in/} which provides books at various high-school levels and in different subjects. No similar openly available resource was found for French and Russian.
    
    \item {\sc Legal}: We identify multiple governmental type of documents that we group under the "legal" domain, which include:\\

    \textbf{Constitutions:} We sample sentences from the U.S. constitution for English, the Lebanese constitution for Arabic, the Indian constitution for Hindi, the French constitution for French, and the Russian constitution for Russian. \\
    
    \textbf{Judicial Rulings:} We used recent public decisions by law courts, such as the Supreme Court in the US \footnote{law.cornell.edu/supremecourt/text}, to collect sentences from judicial rulings, in addition to using legal datasets with such content \cite{judicial-hi}. \\

    \textbf{United Nations Parliament:} We collect samples from the United Nations (UN) Parallel Corpus \cite{un-parallel-corpus} which contains official records and parliamentary documents of the UN. The corpus is available all languages we consider except for Hindi since it is not considered one of the official languages of the UN.
    
    \item {\sc User Reviews}: User text reviews for products, movies, books, hotels, and restaurants, are sampled from open-source datasets in each language when available. Most these datasets are used in sentiment analysis research.
    
    \item {\sc Dialogue}: Conversational text data is collected from three different types of open-source dialogue datasets: \textbf{Open-domain} dialogue datasets which focus on open-ended general conversation \cite{naous-empathetic,open-domain-en,open-domain-hi}, \textbf{Task-oriented} datasets that are design to train human-assistance or customer support dialogue models \cite{task-oriented-en,task-oriented-hi}, and \textbf{Negotiation} dialogues that are used in developing automated sales dialogue agents with negotiation capabilities \cite{negotiation-en}.
    
    \item {\sc Finance}: We leverage the Financial Phrasebank dataset \cite{finance} which provides English sentences with financial references and content collected from finance-focused news, and the CoFiF corpus \cite{finance-fr} which provides financial reports in French.

    \item {\sc Forums}: We collect text from several online forums. These include: \\  
    \textbf{Reddit:} Reddit is a popular platform where online communities discuss common interests and passions. We used the latest version of the Reddit dump available at the time of this study to sample user posts. We filtered posts for language using the fasttext language identification model with a confidence > 0.9. NSFW and Over 18 content were automatically filtered before sampling. Further, any sampled sentence that still contained sexual or offensive content was manually removed.
    
    \textbf{QA Websites:} We collected questions and answers from QA websites using publicly available datasets for Question Answering research \cite{qa-ar,qa-en,hindi-qa,qa-fr,qa-ru}. \\

    \textbf{StackOverflow:} Sentences were collected from the StackOverflow NER dataset \cite{stackoverflow} which contains user posts that describe what the user is trying to accomplish, a problem they are facing, or questions to seek advice from the community.\\

    \item {\sc Social Media}: We sample tweets from the the Stanceosaurus dataset \cite{stanceosaurus} which provides thousands of tweets in English, Arabic, and Hindi that discuss recent region-specific rumors. French tweets were sampled from the dataset of \citet{twitter-fr} built to detect crisis messages in French tweets, while Russian tweets were sampled from the RuSentiTweet dataset \cite{twitter-ru} for sentiment analysis in Russian. Tweets that include offensive or hate speech were manually omitted.

    \item {\sc Policies}: We group under "Policies" several type of documents that delineate plans of what to do in a particular situation. This includes text extracted from: freely available \textbf{contract} templates for apartment/house leasing and job employment, \textbf{Special Olympics rules} which are available in multiple languages among which are but not in Hindi, and online \textbf{codes of conduct} of different organizations that we identify. 
    
    \item {\sc Guides}: Several domains that aim at providing instructions to the reader are grouped under "Guides". We extract data from Samsung Smartphones \textbf{User Manuals} which are available in a variety of languages. Another source is \textbf{Online Tutorials} which we collect from WikiHow that provides how-to articles in multiple languages. We also manually collect \textbf{Recipe Instructions} from multiple online cooking resources for each language.  Additionally, we collect \textbf{Code Documentation} sentences from documentation of different functions of the Matlab software\footnote{mathworks.com}.

    \item {\sc Captions}: We collect four different types of captions: image and video captions from various public datasets used in automatic captioning research, movie subtitles from the OpenSubtitles \cite{open-subtitles} dataset used in machine translation research, and YouTube captions that we manually collect from video released under a Creative Commons license. While high-quality YouTube captions are easy to find for English, we could not find any high-quality YouTube captions for non-English languages.
    
    \item {\sc Medical Text}: We use clinical reports written by medical professionals from the i2b2/VA dataset \cite{clinical-reports-en}. We could not find similar high-quality medical resources for non-English languages.
    
    \item {\sc Dictionaries}: We manually collect sentence examples from Arabic and English dictionaries using words that have appeared in the Word of the Day. No similar resource under a Creative Commons license was found for Hindi, French, and Russian.

    \item {\sc Entertainment}: We use Humour detection datasets to collect jokes \cite{arabic-humour,jokes-en,russian-jokes-dataset}. Hindi jokes were manually collected.
    
    \item {\sc Speech}: Two types of sources for speech data are used: \textbf{publicly available presidential speeches} that are usually posted on governmental websites. We used speeches by the United States President that are posted on the department of state's website. These speeches are also professionally translated to Arabic. We also collect sentences from \textbf{TED Talk transcriptions}, which are professionally translated from English to multiple languages.

\begin{table*}[t]
\begin{adjustbox}{width=\linewidth}
\begin{tabular}{@{}ccl@{}}
\toprule
\multicolumn{1}{l}{\textbf{Lang}} & \textbf{\begin{tabular}[c]{@{}c@{}}Readability\\ Level\end{tabular}} & \multicolumn{1}{c}{\textbf{Distribution (>5\%)}} \\ \midrule
 & \cellcolor[HTML]{EFEFEF}A1 & \cellcolor[HTML]{EFEFEF}Captions (50.62\%) Dialogue (28.4\%) Reviews (7.41\%) \\
 & A2 & Reviews (19.44\%) Dialogue (18.65\%) Guides (17.46\%) Captions (12.7\%) Social Media (5.45\%) Literature (5.95\%) \\
 & \cellcolor[HTML]{EFEFEF}B1 & \cellcolor[HTML]{EFEFEF}Wikipedia (22.37\%) Reviews (15.76\%) Guides (13.23\%) News (10.12\%) Speech (6.03\%) Legal (5.84\%) \\
 & B2 & News (21.59\%) Wikipedia (21.06\%)  Reviews (6.9\%) Entertainment (6.73\%) Legal (6.55\%) Policies (6.37\%) Speech (5.31\%) \\
 & \cellcolor[HTML]{EFEFEF}C1 & \cellcolor[HTML]{EFEFEF}Wikipedia (40.29\%) Research (14.53\%) Literature (13.43\%) Textbooks (5.71\%) \\
\multirow{-6}{*}{\textbf{ar}} & C2 & Poetry (24.04\%) Wikipedia (26.23\%) Novels (18.58\%) Dictionaries (9.84\%) Quotes (6.01\%) \\ \midrule
 & \cellcolor[HTML]{EFEFEF}A1 & \cellcolor[HTML]{EFEFEF}Captions (44.29\%) Dialogue (9.29\%) Twitter (8.57\%) Poetry (7.86\%) Quotes (5\%) \\
 & A2 & Recipes (9.02\%) Dialogue (12.02\%) Twitter (7.1\%) Quotes (7.1\%) QA Websites (6.28\%) Children Stories (5.46\%) \\
 & \cellcolor[HTML]{EFEFEF}B1 & \cellcolor[HTML]{EFEFEF}Wikipedia (21.85\%) Guides (15.32\%) Books (10.36\%) Legal (6.98\%) Reddit (5.41\%) \\
 & B2 & Wikipedia (43.47\%) Legal (10.51\%) Policies (9.66\%) Books (7.39\%) Guides (6.25\%) \\
 & \cellcolor[HTML]{EFEFEF}C1 & \cellcolor[HTML]{EFEFEF}Wikipedia (46.47\%) Policies (12.03\%) Research (9.96\%) Finance (7.74\%) \\
\multirow{-6}{*}{\textbf{fr}} & C2 & Research (21.43\%) Policies (7.14\%) Finance (6.39\%) \\ \midrule
 & \cellcolor[HTML]{EFEFEF}A1 & \cellcolor[HTML]{EFEFEF}Dialogue (38.25\%) Captions (27.87\%) Reviews (10.38\%) Guides (5.46\%) \\
 & A2 & Captions (16.74\%)  Reviews (13.33\%) Statements (8.15\%) Guides (10.03\%) Dialogue (8.74\%) Forums (7.41\%) Entertainment (5.63\%) \\
 & \cellcolor[HTML]{EFEFEF}B1 & \cellcolor[HTML]{EFEFEF}Wikipedia (16.72\%) Reviews (13.85\%) News (11.74\%) Forums (7.8\%) Guides (8.12\%) Textbooks (7.17\%) \\
 & B2 & Wikipedia (21.94\%) News (11.8\%) Research (10.8\%) Textbooks (11.03\%) Policies (7.83\%) Literature (7.39\%) \\
 & \cellcolor[HTML]{EFEFEF}C1 & \cellcolor[HTML]{EFEFEF}Wikipedia (24.23\%) Research (13.14\%) Literature (12.82\%) Legal (9.54\%)  Textbooks (9.28\%) Policies (5.67\%) News (5.65\%) \\
\multirow{-6}{*}{\textbf{en}} & C2 & Wiki-Natural Sciences (16.25\%) Literature (18.75\%) Clinical Reports (11.25\%) Research (8.7\%) Textbooks (7.5\%) \\ \midrule
 & \cellcolor[HTML]{EFEFEF}A1 & \cellcolor[HTML]{EFEFEF}Captions (33.09\%) Literature (16.91\%) Dialogue (12.82\%) Jokes (9.56\%) Reviews (5.15\%) \\
 & A2 & Captions (12.88\%) Dialogue (12.88\%)  Forums (7.46\%)  Statements (7.46\%) Children Stories (6.78\%)  (5.37\%) Guides (5.76\%) \\
 & \cellcolor[HTML]{EFEFEF}B1 & \cellcolor[HTML]{EFEFEF}Wikipedia (15.02\%) Literature (13.31\%) Guides (11.26\%) Reviews (9.56\%) Statements (8.53\%) Forums (8.53\%) \\
 & B2 & Wikipedia (21.27\%) Textbooks (9.7\%) Literature (9.33\%) Poetry (8.96\%) Research (7.46\%) Policies (7.46\%) Quotes (5.6\%) \\
 & \cellcolor[HTML]{EFEFEF}C1 & \cellcolor[HTML]{EFEFEF}Wikipedia (31.08\%) Textbooks (12.16\%) Legal (10.36\%) Research (10.36\%)  Literature (8.53\%) Forums (7.21\%) Poetry (5.41\%) \\
\multirow{-6}{*}{\textbf{hi}} & C2 & Wikipedia (44.25\%)  Textbooks (10.92\%) Legal (10.9\%)  Research (8.05\%) \\ \midrule
 & \cellcolor[HTML]{EFEFEF}A1 & \cellcolor[HTML]{EFEFEF}Reviews (10.7\%) Recipes (9.2\%) Twitter (9.45\%) Dialogue (8.21\%) Jokes (7.96\%) Captions (5.97\%) \\
 & A2 & Wikipedia (23.80\%) Guides (15.36\%) Research (8.19\%) Speech (7.14\%) \\
 & \cellcolor[HTML]{EFEFEF}B1 & \cellcolor[HTML]{EFEFEF}Wikipedia (32.76\%) Guides (6.11\%) Policies (5.62\%) Legal (5.62\%) \\
 & B2 & Wikipedia (34.05\%) Research (20.86\%) Legal (12.88\%) Policies (9.51\%) Community Websites (6.13\%) \\
 & \cellcolor[HTML]{EFEFEF}C1 & \cellcolor[HTML]{EFEFEF}Wikipedia (31.65\%) Research (26.16\%) Legal (19.38\%) Policies (8.81\%) \\
\multirow{-6}{*}{\textbf{ru}} & C2 & Legal (28.42\%) Research (17.58\%) Policies (6.59\%) \\ \bottomrule
\end{tabular}
\end{adjustbox}
\caption{Distribution of domains for each readability level in each language. Only domains that compose more than 5\% of the distribution are show.}
\label{tab:dist-domains}
\end{table*}

    \item {\sc Statements}: Two different types of standalone sentences that we group under "statements" were identified which are: Rumours, and quotes. We collect rumours in Arabic, English, and Hindi from the Stanceosaurus dataset \cite{stanceosaurus} used in misinformation detection. The rumours/claims are collected from various fact-checking websites in the Arab World, India, and the U.S. We also manually collected quotes in the three languages from various online resources. We did not collect mere translations of famous English quotes to other languages but focused on quotes by old scholars and thinkers of the Arab World, France, Russia and India for more cultural representation.
    
    \item {\sc Poetry}: Poetry lines are extracted from English, Arabic, and Hindi poems, some of which date back several centuries ago. To have culture specific samples, we focus on non-English poems from original Arab, French, Indian, and Russian authors, and not poems translated from English.
    
    \item {\sc Letters}: English letters were collected from  online archives of historic letters. No high-quality authentic letters were found in Arabic or Hindi.

\end{itemize}

\subsection{Domain Distribution}
\begin{figure}[h!]
    \centering
    \includegraphics[width=0.85\linewidth]{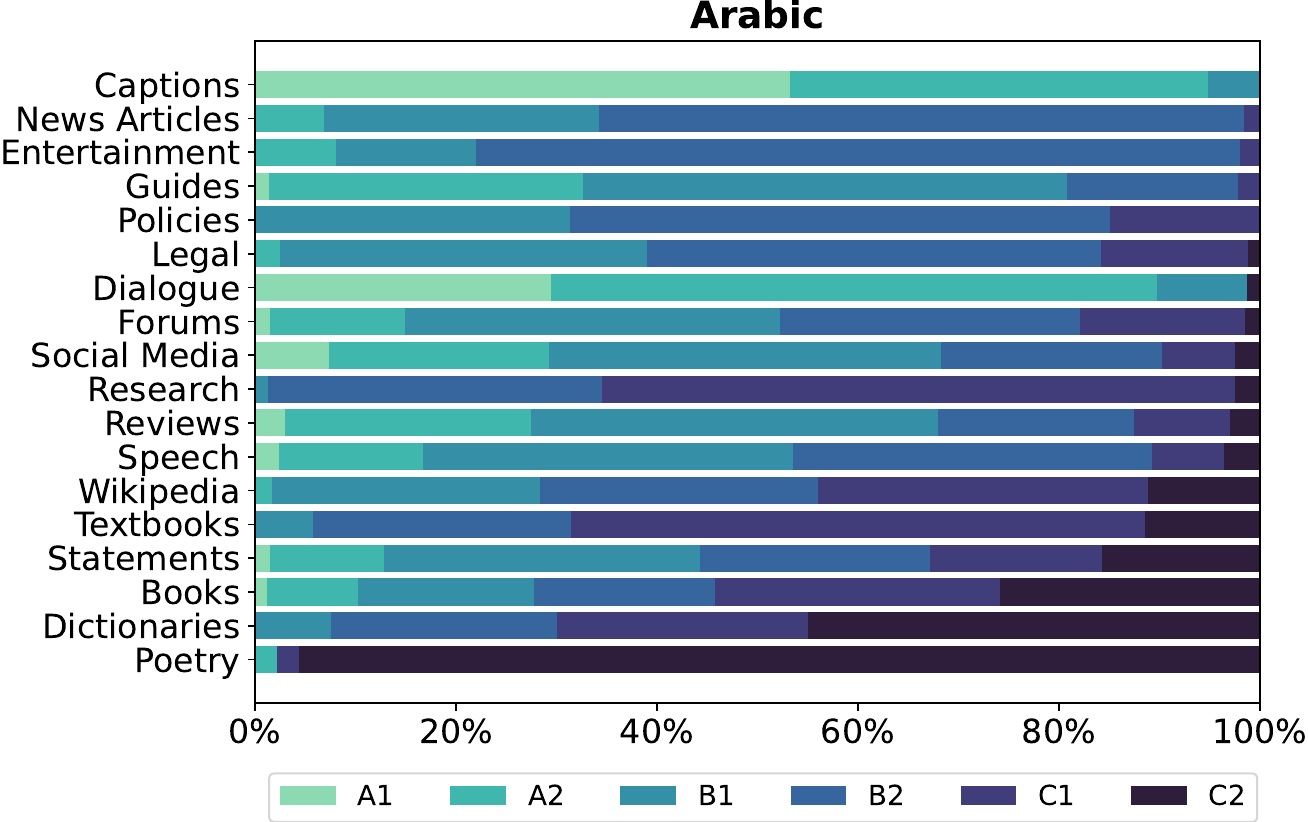}
    \includegraphics[width=0.85\linewidth]{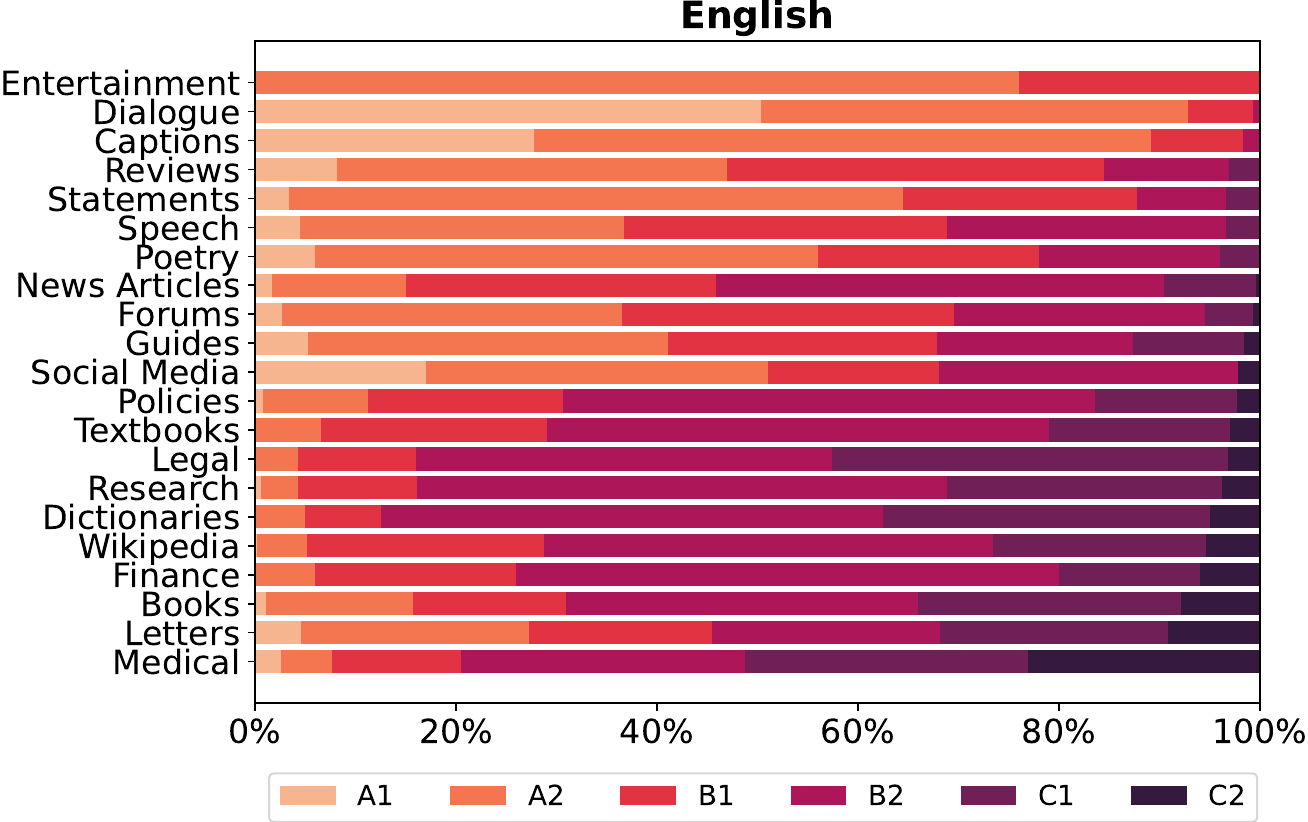}
    \includegraphics[width=0.85\linewidth]{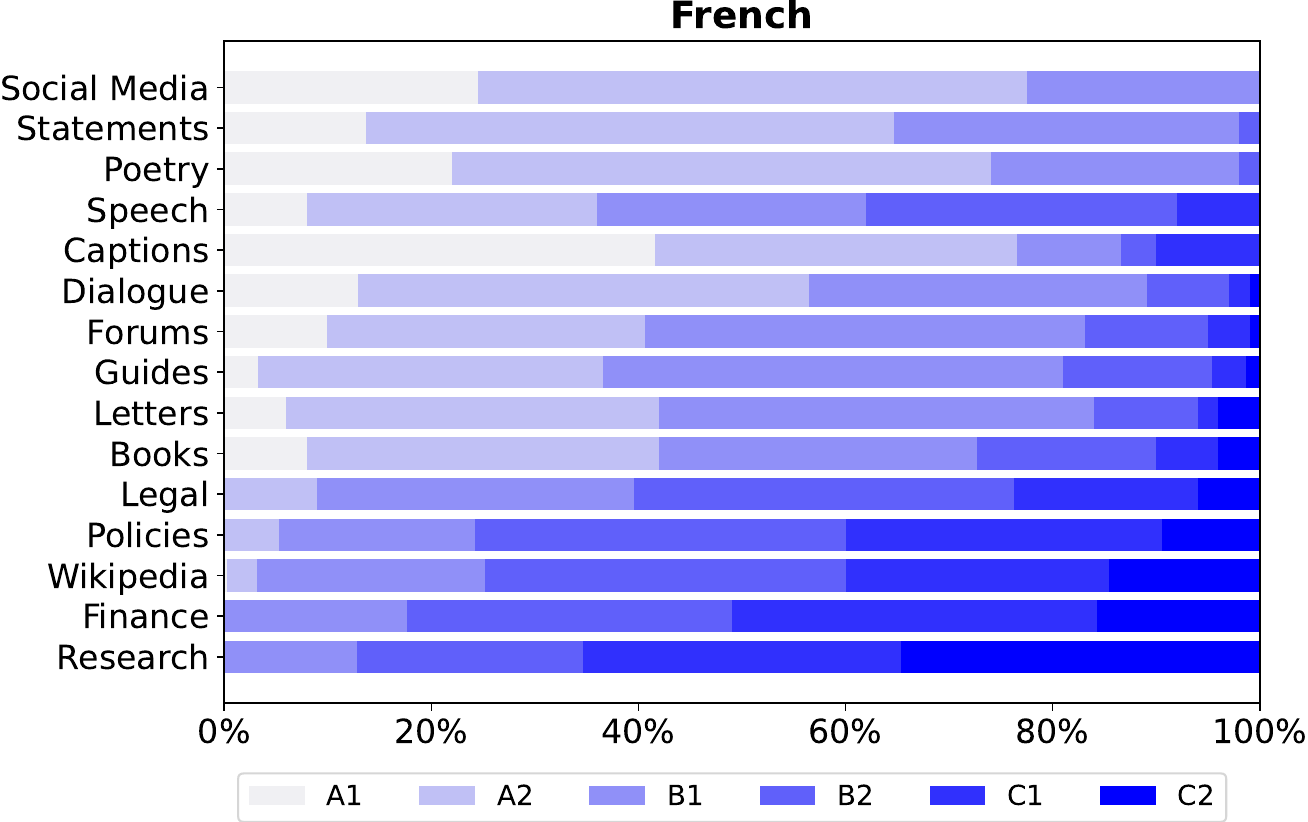}
    \includegraphics[width=0.85\linewidth]{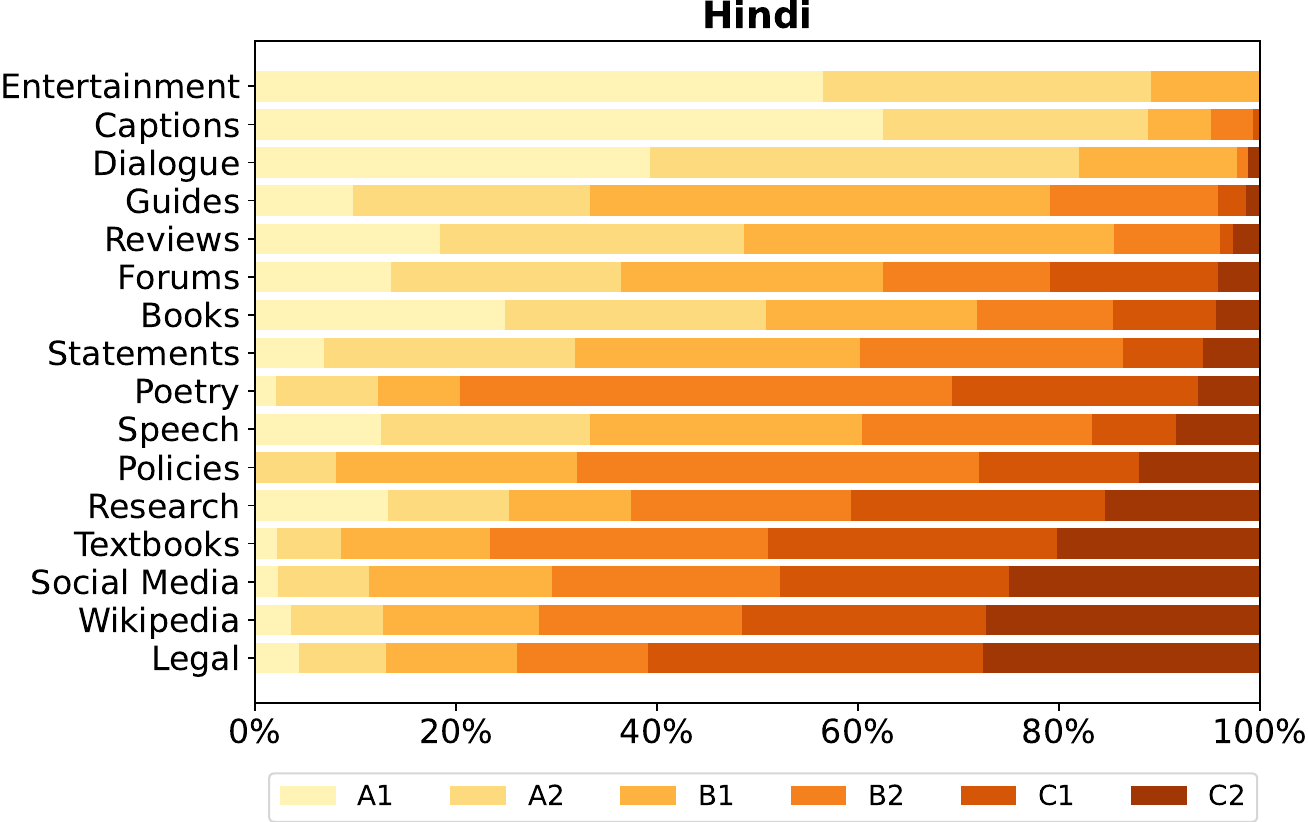}
    \includegraphics[width=0.85\linewidth]{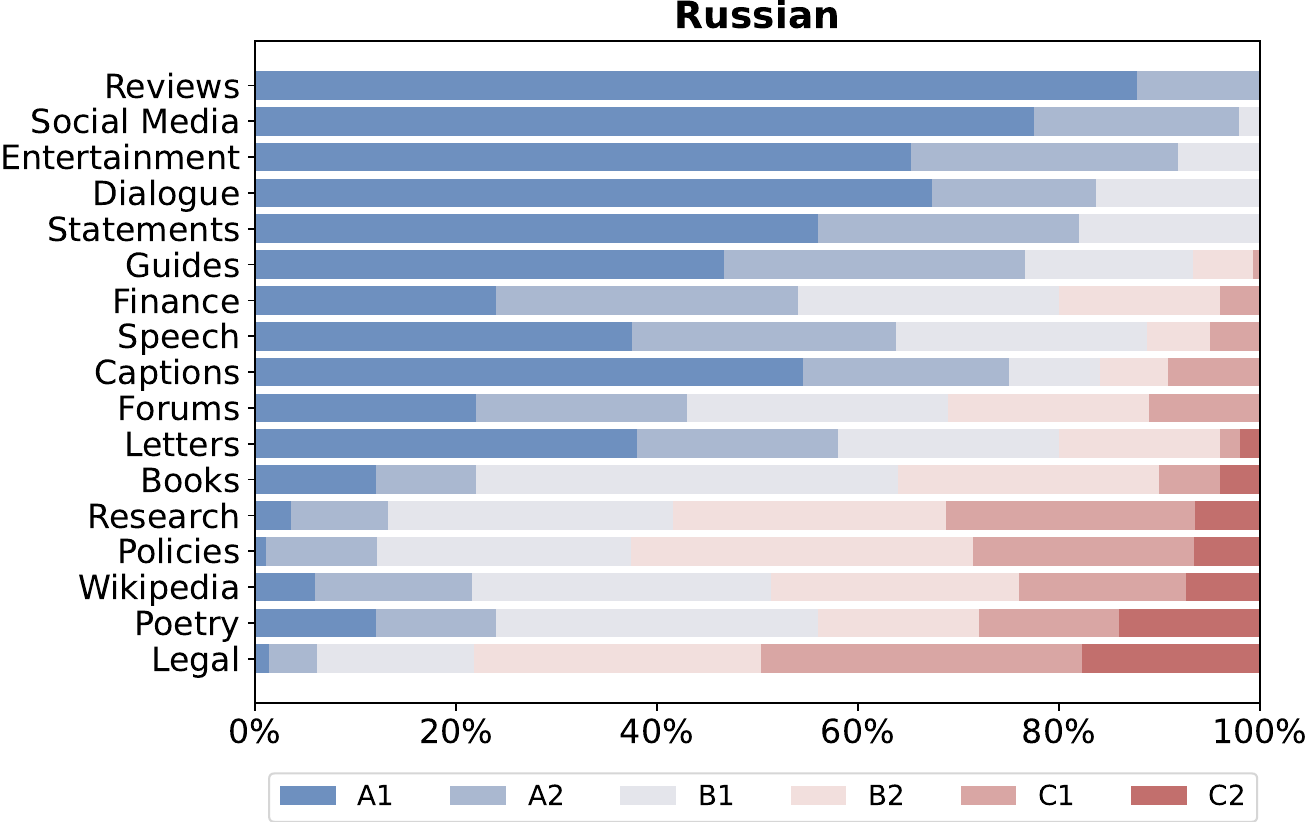}
    \caption{The readability levels vary greatly across domains and languages in {\sc ReadMe++}, highlighting the importance to consider diversity of data sources.}
    \label{fig:per-domain}
\end{figure}

Table~\ref{tab:dist-domains} shows the distribution of the domains in each readability level for each language. Basic readability levels (A1, A2) mostly contains sentences from domains that have text that is straightforward to read and contains day-to-day vocabulary such as Captions, Dialogue, User Reviews, User Guides. Intermediate readability levels (B1, B2) largely contain sentences from domains that present factual content such as books, Wikipedia articles, policy documents, news articles, etc. Proficient levels (C1, C2) contain domains that are scientific and technical such as finance, medical, legal documents, or highly literary text such as Arabic Poetry. We show the distribution of readability levels per domain in Figure~\ref{fig:per-domain}.

\subsection{Sentence Examples}
    
\begin{figure*}[h]
    \centering
    \includegraphics{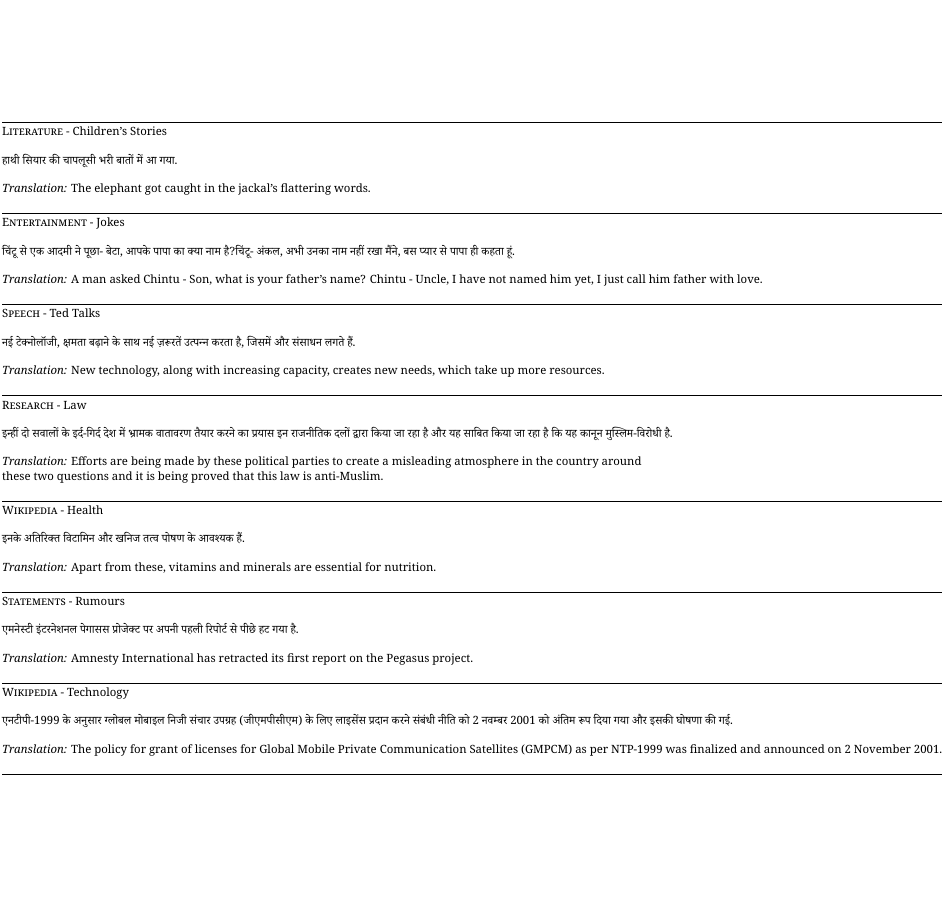}
    \caption{Hindi sentence examples from {\sc ReadMe++}.}
    \label{fig:hindi-examples}
\end{figure*}

\begin{figure*}[h]
    \centering
    \includegraphics{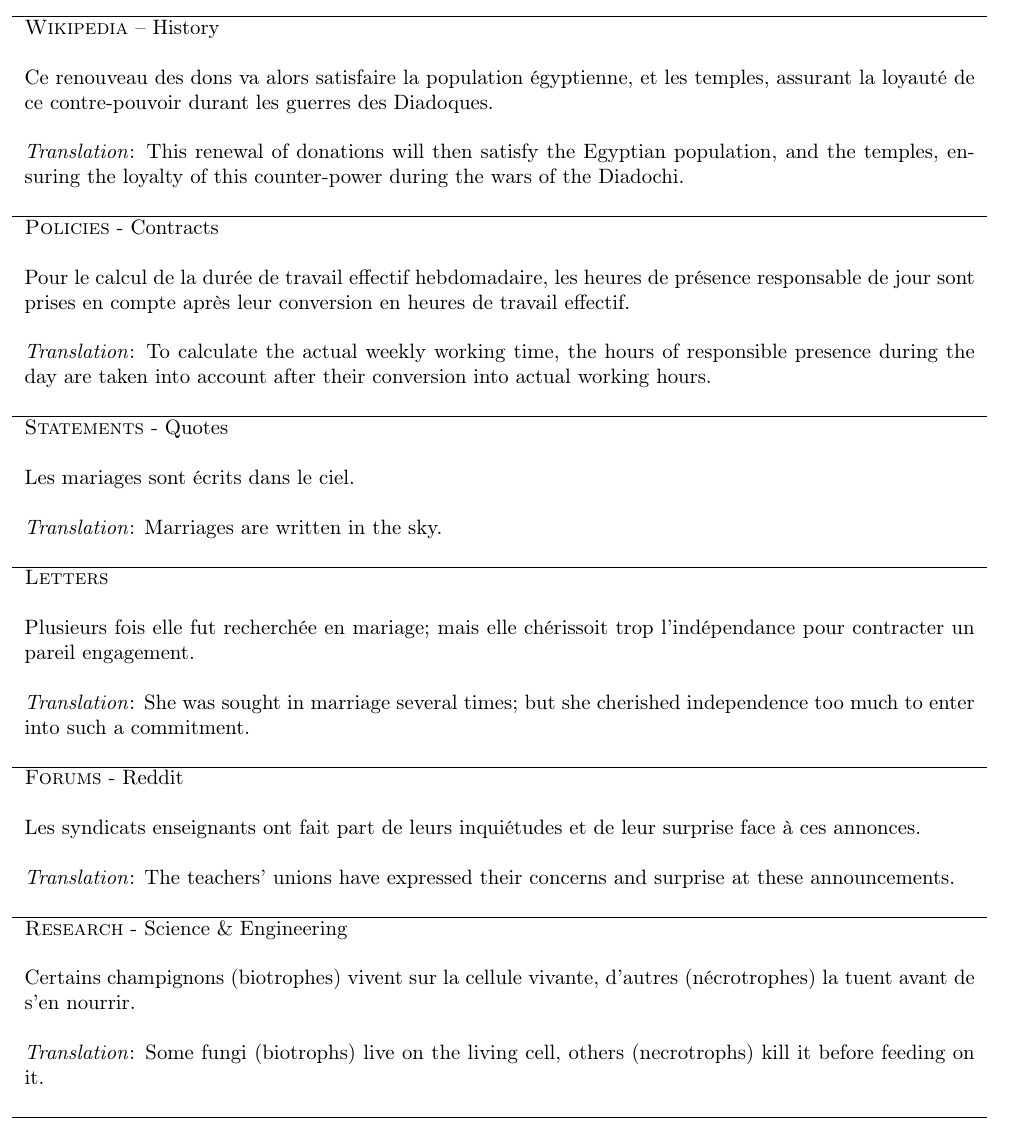}
    \caption{French sentence examples from {\sc ReadMe++}.}
    \label{fig:examples-fr}
\end{figure*}

\begin{figure*}[h]
    \centering
    \includegraphics{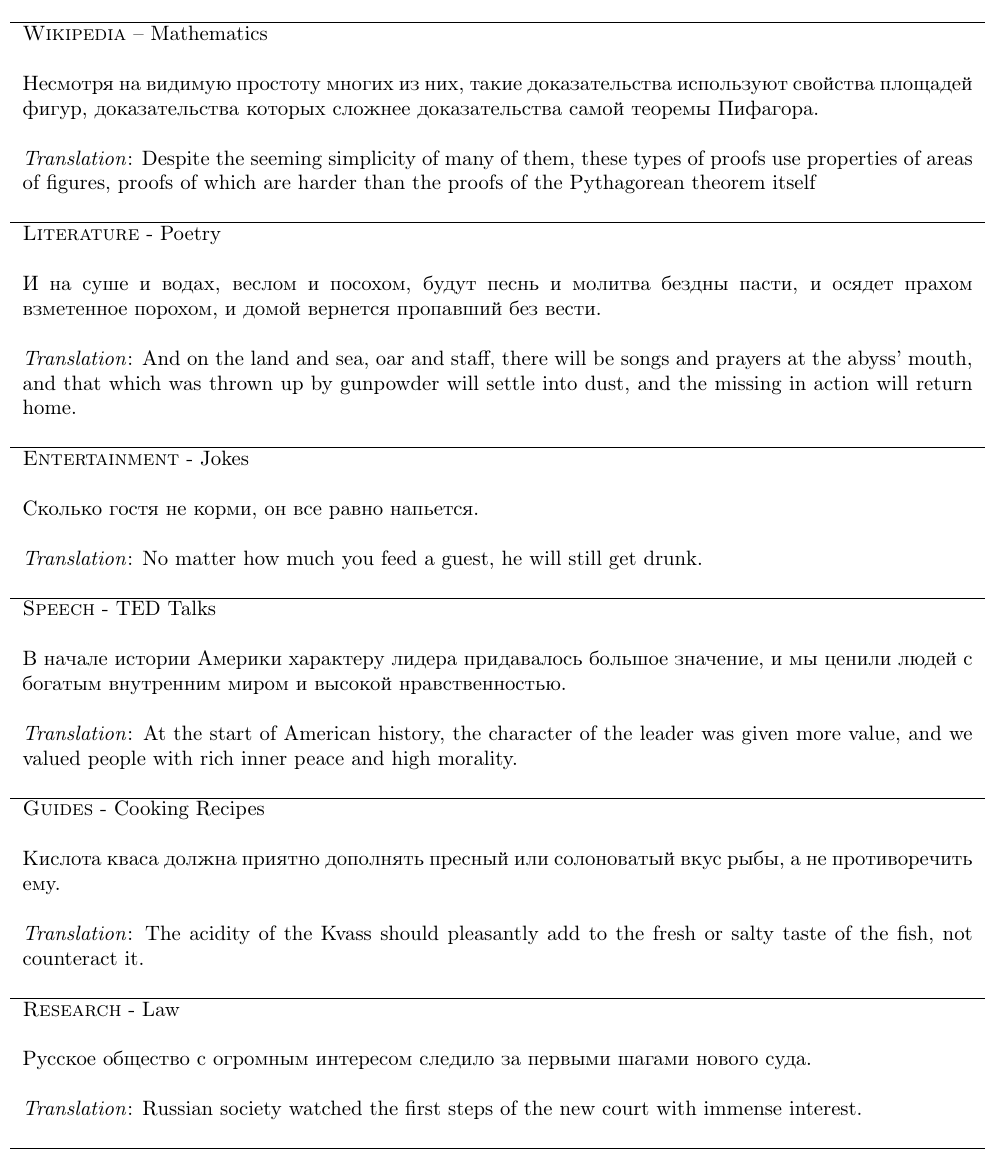}
    \caption{Russian sentence examples from {\sc ReadMe++}.}
    \label{fig:examples-ru}
\end{figure*}

Example sentences from various domains are shown in Table~\ref{tab:examples-en} for English, Table~\ref{tab:examples-ar} for Arabic, 
Figure~\ref{fig:hindi-examples} for Hindi,
Figure~\ref{fig:examples-fr} for French,  and Figure~\ref{fig:examples-ru} for Russian.

\section{CEFR Levels Descriptors}
\label{app:cefr-levels}

The CEFR levels descriptors are provided in Table~\ref{tab:cefr-descriptions}. Each level is described by specific capabilities of a language learner which we used to familiarize annotators with the intuition behind the scale being used prior to labeling.

    \begin{table*}[t]
\begin{adjustbox}{width=\linewidth}
\begin{tabular}{@{}ll@{}}
\toprule
\textbf{CEFR Level} & \textbf{Description} \\ \midrule
\textbf{A1} & \begin{tabular}[c]{@{}l@{}}Can understand and use familiar everyday expressions and very basic phrases aimed at the satisfaction of needs of a concrete type.\\ Can introduce him/herself and others and can ask and answer questions about personal details such as where he/she lives, people he/she knows and things he/she has.\\ Can interact in a simple way provided the other person talks slowly and clearly and is prepared to help.\end{tabular} \\ \midrule
\textbf{A2} & \begin{tabular}[c]{@{}l@{}}Can understand sentences and frequently used expressions related to areas of most immediate relevance (e.g. basic personal information, employment, etc.). \\ Can communicate in simple and routine tasks requiring a simple and direct exchange of information on familiar and routine matters.\\ Can describe in simple terms aspects of his/her background, immediate environment and matters in areas of immediate need.\end{tabular} \\ \midrule
\textbf{B1} & \begin{tabular}[c]{@{}l@{}}Can understand the main points of clear standard input on familiar matters regularly encountered in work, school, leisure, etc.\\ Can deal with most situations likely to arise whilst travelling in an area where the language is spoken.\\ Can produce simple connected text on topics which are familiar or of personal interest.\\ Can describe experiences and events, dreams, hopes and ambitions and briefly give reasons and explanations for opinions and plans.\end{tabular} \\ \midrule
\textbf{B2} & \begin{tabular}[c]{@{}l@{}}Can understand the main ideas of complex text on both concrete and abstract topics, including technical discussions in his/her field of specialisation.\\ Can interact with a degree of fluency and spontaneity that makes regular interaction with native speakers quite possible without strain for either party. \\ Can produce clear, detailed text on a wide range of subjects and explain a viewpoint on a topical issue giving the advantages and disadvantages of various options.\end{tabular} \\ \midrule
\textbf{C1} & \begin{tabular}[c]{@{}l@{}}Can understand a wide range of demanding, longer texts, and recognise implicit meaning.\\ Can express him/herself fluently and spontaneously without much obvious searching for expressions.\\ Can use language flexibly and effectively for social, academic and professional purposes.\\ Can produce clear, well-structured, detailed text on complex subjects, showing controlled use of organisational patterns, connectors and cohesive devices.\end{tabular} \\ \midrule
\textbf{C2} & \begin{tabular}[c]{@{}l@{}}Can understand with ease virtually everything heard or read. \\ Can summarise information from different spoken and written sources, reconstructing arguments and accounts in a coherent presentation. \\ Can express him/herself spontaneously, very fluently and precisely, differentiating finer shades of meaning even in more complex situations.\end{tabular} \\ \bottomrule
\end{tabular}
\end{adjustbox}
\caption{Level descriptions of the CEFR scale used for readability annotation.}
\label{tab:cefr-descriptions}
\end{table*}

\newpage
\section{Traditional Metrics}
\label{appendix:feature-based-metrics}

ARI and FKGL are statistical formulas  based on the number of words, characters, and syllables.

\paragraph{Automated Readability Index (ARI).} ARI aims at approximating the grade level needed by an individual to understand a text. It is computed by:

\begin{equation}
\small{
   \mathrm{ARI} = 4.71 \left(\frac{ \mathrm{\#Chars}}{\mathrm{\#Words}}\right) + 0.5\left(\frac{\mathrm{\#Words}}{\mathrm{\#Sents}}\right) -21.43
    }
\end{equation}

\paragraph{Flesch-Kincaid Grade Level (FKGL).} FKGL also aims at predicting the grade level, but unlike ARI, considers the total number of syllables in the text. It is computed as follows: 

\begin{equation}
\small{
   \mathrm{FKGL} = 0.39 \left(\frac{ \mathrm{\#Words}}{\mathrm{\#Sents}}\right) + 11.8\left(\frac{\mathrm{\#Sylla}}{\mathrm{\#Words}}\right) -15.59
    }
\end{equation}

\paragraph{Open Source Metric for Measuring Arabic Narratives (OSMAN).} OSMAN is computed according to the following formula:

\begin{equation}
\begin{split}
   \mathrm{OSMAN} = 200.791 -1.015 \left(\frac{A}{B}\right) +  \\
   24.181 \left( \frac{C}{A} + \frac{D}{A} + \frac{G}{A} + \frac{H}{A}  \right)
\end{split}
\end{equation}

where \textit{A} is the number of words, \textit{B} is the number of sentences, \textit{C} is the number of words with more than 5 letters, \textit{D} is the number of syllables, \textit{G} is the number of words with more than four syllabus, and \textit{H} is the number of "Faseeh" words, which contain any of the letters
(\setcode{utf8}\<ظ، ذ، ؤ، ئ، ء>)
or end with 
(\setcode{utf8}\<ون، وا>).

\begin{table}[h!]
\centering
\begin{adjustbox}{width=\linewidth}
\begin{tabular}{lcllll}
\toprule
\multicolumn{1}{c}{} &  & \multicolumn{4}{c}{\textbf{Pre-training Sources}} \\ \cline{3-6} 
\multicolumn{1}{c}{\multirow{-2}{*}{\textbf{Model}}} & \multirow{-2}{*}{\textbf{\#Params}} & \multicolumn{1}{c}{Wiki} & \multicolumn{1}{c}{News} & \multicolumn{1}{c}{Books} & \multicolumn{1}{c}{CC} \\ \midrule
\multicolumn{6}{l}{\cellcolor[HTML]{EFEFEF}\small{Multilingual LMs}} \\
mBERT & 177M & $\checkmark$ &  &  &  \\
XLMR$_{B}$ & 278M &  &  &  & $\checkmark$ \\
XLMR$_{L}$ & 559M &  &  &  & $\checkmark$ \\
mT5$_{S}$ & 60M &  &  &  & $\checkmark$ \\
mT5$_{B}$ & 220M &  &  &  & $\checkmark$ \\
mT5$_{L}$ & 770M &  &  &  & $\checkmark$ \\
Aya101       & 13B  &    &  &  &  $\checkmark$ \\
\multicolumn{6}{l}{\cellcolor[HTML]{EFEFEF}\small{Monolingual Arabic LMs}} \\
AraBERT$_{B}$ & 135M & $\checkmark$ & $\checkmark$ &  &  \\
AraBERT$_{L}$ & 369M & $\checkmark$ & $\checkmark$ &  & $\checkmark$ \\
ArBERT & 163M & $\checkmark$ & $\checkmark$ & $\checkmark$ & $\checkmark$ \\
AraT5$_{B}$ & 220M & $\checkmark$  & $\checkmark$ &  $\checkmark$ & $\checkmark$ \\
\multicolumn{6}{l}{\cellcolor[HTML]
{EFEFEF}\small{Monolingual French LMs}} \\
CamemBERT$_{B}$ & 110M &   &   &  & $\checkmark$  \\
CamemBERT$_{L}$ & 335M &   &   &  & $\checkmark$   \\
\multicolumn{6}{l}{\cellcolor[HTML]
{EFEFEF}\small{Monolingual English LMs}} \\
BERT$_{B}$ & 110M & $\checkmark$ &  & $\checkmark$ &  \\
BERT$_{L}$ & 350M & $\checkmark$ &  & $\checkmark$ &  \\
\multicolumn{6}{l}{\cellcolor[HTML]{EFEFEF}\small{Indian LMs}} \\
MuRIL$_{B}$ & 237M & $\checkmark$ &  &  & $\checkmark$ \\
MuRIL$_{L}$ & 506M & $\checkmark$ & \multicolumn{1}{c}{} & \multicolumn{1}{c}{} & $\checkmark$ \\
IndicBERTv2$_{B}$ & 278M &    & $\checkmark$ &   & $\checkmark$ \\
\multicolumn{6}{l}{\cellcolor[HTML]
{EFEFEF}\small{Monolingual Russian LMs}} \\
RuBERT$_{B}$ & 180M & $\checkmark$    &  &   &  \\
\bottomrule
\end{tabular}
\end{adjustbox}
\caption{Summary of LMs used in experiments. \textbf{CC} stands for Common Crawl. }
\label{tab:summary-models}
\end{table}

\section{Experimental Details}

\subsection{Language Models}
\label{appendix:pretrained-models}

The details of the pre-trained LMs used in our experiments are provided in Table~\ref{tab:summary-models}, including the number of parameters and pre-training data sources. The majority of models have been pre-trained using CommonCrawl data. Aya is based on mT5$_{XXL}$ and further instruction-tuned using the Aya dataset \cite{singh2024aya}. Training was performed using four NVIDIA A40 GPUs. We fine-tuned Aya using LoRA \cite{hu2021lora} and 4-bit quantization. We set LoRa hyperparameters as follows: rank=8, alpha=16, dropout=0.05.

\subsection{Corpus Split}
\label{app:split-stats}

The train/validation/test split statistics of {\sc ReadMe++} are shown in Table~\ref{tab:split-statistics} for each language. Those splits are obtained based on taking a 60\%/10\%/30\% split for train/validation/test per domain, ensuring all domains are covered in each split.

\begin{table}[t]
\begin{adjustbox}{width=\linewidth}
\begin{tabular}{@{}ccccccccc@{}}
\toprule
\multirow{2}{*}{\textbf{Lang}} & \multirow{2}{*}{\textbf{Split}} & \multicolumn{7}{c}{\textbf{Readability Class}} \\ \cmidrule(l){3-9} 
 &  & 1$_{(A1)}$ & 2$_{(A2)}$ & 3$_{(B1)}$ & 4$_{(B2)}$ & 5$_{(C1)}$ & 6$_{(C2)}$ & Total \\ \midrule
\multirow{3}{*}{\textbf{ar}} & \#train & 49 & 151 & 307 & 324 & 207 & 114 & 1152 \\
 & \#val & 6 & 25 & 53 & 62 & 35 & 17 & 198 \\
 & \#test & 26 & 76 & 154 & 179 & 108 & 52 & 595 \\ \midrule
 \multirow{3}{*}{\textbf{fr}} & \#train & 78 & 226 & 270 & 200 & 144 & 72 & 990 \\
 & \#val & 13 & 35 & 34 & 44 & 22 & 15 & 163 \\
 & \#test & 49 & 105 & 140 & 108 & 75 & 39 & 516 \\ \midrule
\multirow{3}{*}{\textbf{en}} & \#train & 105 & 414 & 354 & 536 & 245 & 49 & 1703 \\
 & \#val & 20 & 61 & 64 & 99 & 30 & 8 & 282 \\
 & \#test & 58 & 200 & 210 & 272 & 113 & 23 & 876 \\ \midrule
\multirow{3}{*}{\textbf{hi}} & \#train & 158 & 182 & 170 & 148 & 121 & 118 & 897 \\
 & \#val & 29 & 27 & 27 & 28 & 29 & 12 & 152 \\
 & \#test & 85 & 86 & 96 & 92 & 72 & 44 & 475 \\ \midrule
 \multirow{3}{*}{\textbf{ru}} & \#train & 235 & 174 & 252 & 191 & 151 & 49 & 1052 \\
 & \#val & 42 & 23 & 42 & 35 & 20 & 13 & 175 \\
 & \#test & 125 & 96 & 115 & 100 & 66 & 29 & 531 \\ 
 \bottomrule
\end{tabular}
\end{adjustbox}
\caption{Number of sentences per readability level for each data split of {\sc ReadMe++}.}
\label{tab:split-statistics}
\end{table}


\subsection{Few-shot Prompt}
\label{app:prompting}
The prompt used for GPT3.5, GPT4, and Llama-7B is provided in Table \ref{tab:llama-prompt}.  The prompt contains 5 primary parts: The task description, definition of readability, example CEFR levels, example sentences with readability scores, and finally the new sentence for evaluation.  When investigating the importance of the few-shot demonstrations we modified how we sampled the few-shot examples from the training set, however the prompt scaffolding remained the same.

\begin{table*}[]
\centering
\begin{adjustbox}{width=1\textwidth}
\begin{tabular}{@{}l@{}}
\toprule
\begin{tabular}[c]{@{}l@{}}

\texttt{Rate the following sentence on it's readability level.  The readabilty is defined} \\ \texttt{as the cognitive load required to understand the meaning of the sentence.  Rate} \\ \texttt{the readabilty on a scale from very easy to very hard.  Base your scores off the}\\ \texttt{CEFR scale for L2 Learners.  You should use the following key:}\\
\\
\texttt{1 = Can understand very short, simple texts a single phrase at a time, picking up} \\ \texttt{familiar names, words and basic phrases and rereading as required.}\\
\texttt{2 = Can understand short, simple texts on familiar matters of a concrete type}\\
\texttt{3 = Can read straightforward factual texts on subjects related to his/her field} \\ \texttt{and interest with a satisfactory level of comprehension.}\\
\texttt{4 = Can read with a large degree of independence, adapting style and speed of} \\ \texttt{reading to different texts and purpose}\\
\texttt{5 = Can understand in detail lengthy, complex texts, whether or not they relate} \\ \texttt{to his/her own area of speciality, provided he/she can reread difficult sections.}\\
\texttt{6 = Can understand and interpret critically virtually all forms of the written} \\ \texttt{language including abstract, structurally complex, or highly colloquial literary}\\ \texttt{and non-literary writings.} \\
\\
\\
\texttt{EXAMPLES:}\\
\texttt{Sentence: "[EX 1]"}\\
\texttt{Given the above key, the readability of the sentence is (scale=1-6): [EX RATING 1]}\\ \\

\texttt{Sentence: "[EX 2]"}\\
\texttt{Given the above key, the readability of the sentence is (scale=1-6): [EX RATING 2]} \\ \\

\texttt{...} \\ \\

\texttt{Sentence: "[EX N]"}\\
\texttt{Given the above key, the readability of the sentence is (scale=1-6): [EX RATING N]} \\ \\

\texttt{Sentence: "[SENTENCE]"}\\
\texttt{Given the above key, the readability of the sentence is (scale=1-6):} 

\end{tabular} \\
 \\ \bottomrule
\end{tabular}
\end{adjustbox}
\caption{Prompt provided to GPT4, GPT3.5, Aya23-8b, Llama2-7b, and Llama3.1-8b models to assess in-context learning readability assessment capabilities.}
\label{tab:llama-prompt}
\end{table*}

\newpage
\section{Additional Results}

\subsection{Main Results: Additional Metrics}
\label{app:main-f1-scores}

\begin{figure*}[t]
    \centering
    \includegraphics[width=\linewidth]{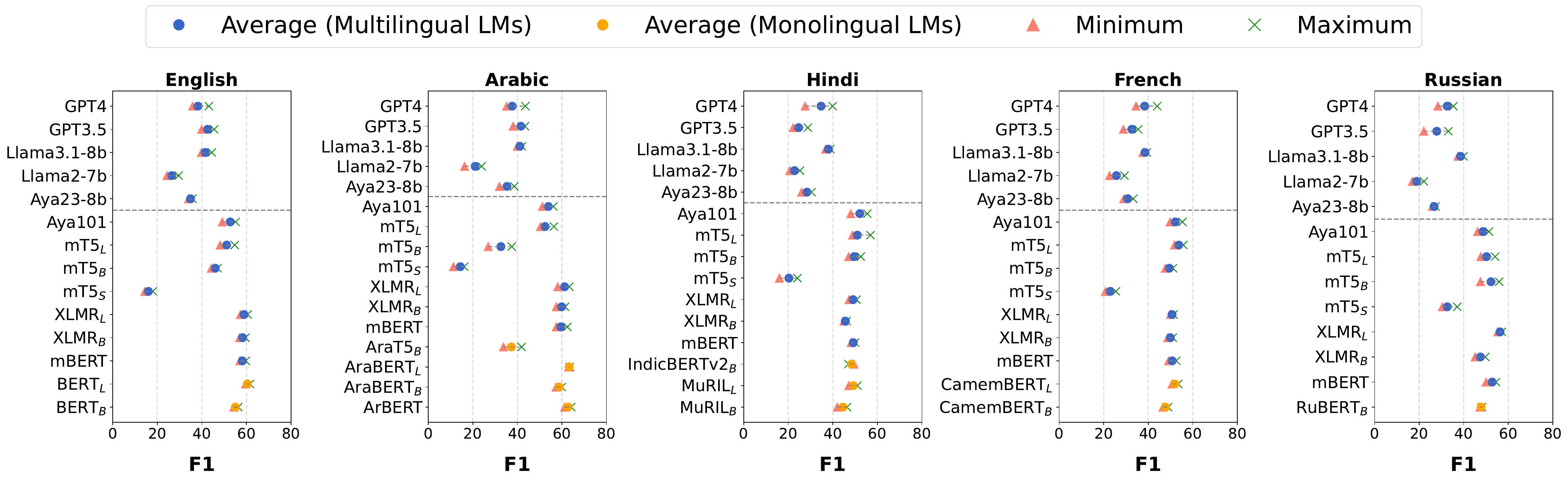}
    \caption{F1 score results of supervised fine-tuning and few-shot prompting on the test set of {\sc ReadMe++}.}
    \label{fig:sup-results-F1}
\end{figure*}

\begin{figure*}[t]
    \centering
    \includegraphics[width=\linewidth]{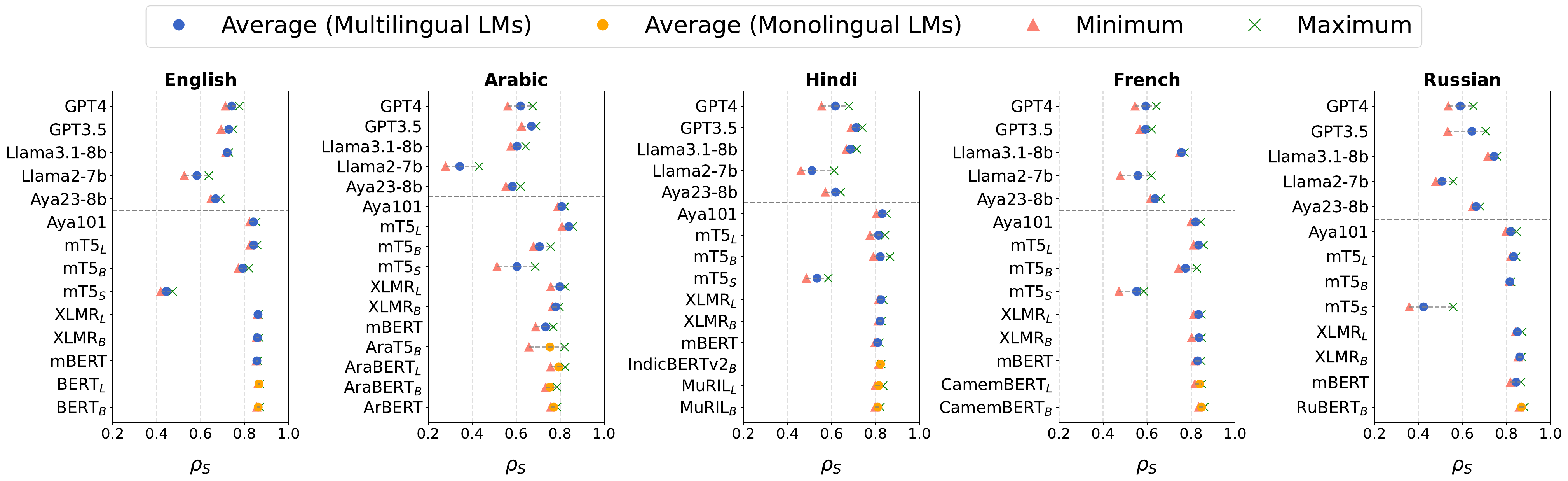}
    \caption{Spearman Correlation ($\rho_{S}$) of supervised fine-tuning and few-shot prompting on the test set of {\sc ReadMe++}.}
    \label{fig:sup-results-spearman}
\end{figure*}

The F1 scores obtained by the fine-tuned models are shown in Figure~\ref{fig:sup-results-F1}. We also report the Spearman Correlation ($\rho_{S}$) as an additional correlation measure in Figure~\ref{fig:sup-results-spearman}. The same trends for models observed in \S\ref{sec:supervised-methods} hold for other metrics.





\subsection{Domain Correlation}
\label{app:domain-correlation}

\begin{figure*}[t!]
    \centering
    \includegraphics[width=\linewidth]{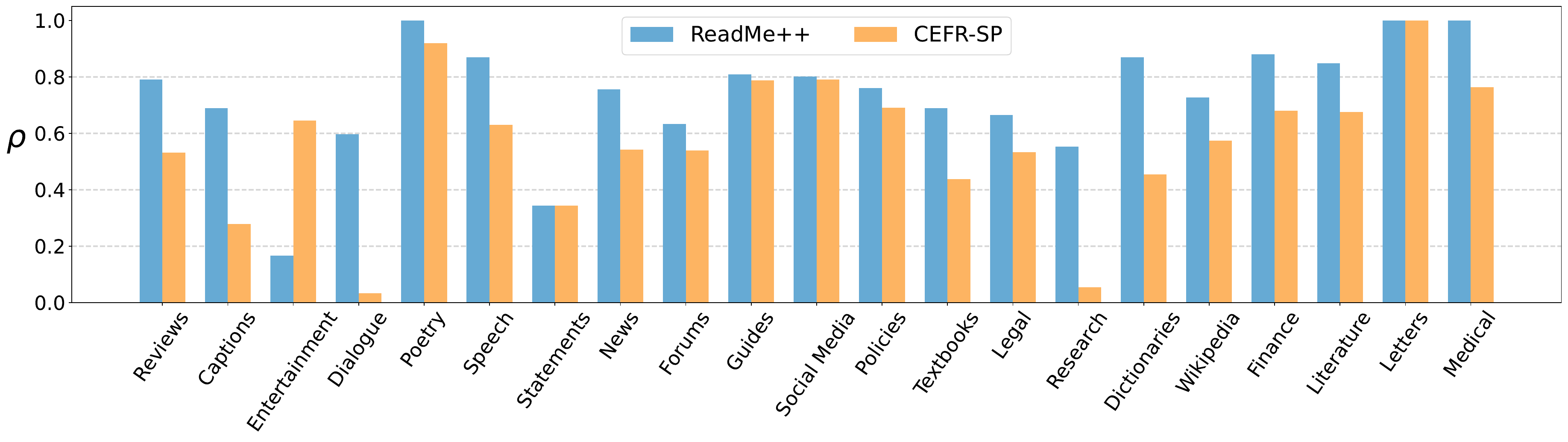}
    \caption{Pearson Correlation per domain for XLMR$_{L}$ trained using {\sc ReadMe++} and CEFR-SP. The model trained with {\sc ReadMe++} achieves better domain generalization, shown by higher correlation in all but one domain (Entertainment).}  
    \label{fig:domain-correlation}
\end{figure*}

To explore the utility of the large data diversity in {\sc ReadMe++}, we investigate the performance of models trained on both {\sc ReadMe++} and CEFR-SP across several specific domains.  We train XLMR$_{L}$ using the publicly available Wikipedia splits of CEFR-SP (1 data source) compared to the public data from {\sc ReadMe++} (112 data sources) The correlation of model predictions with human annotated labels are shown for 21 different textual domains in Figure \ref{fig:domain-correlation}.  In 18 out of the 21 domains, the model trained on {\sc ReadMe++} clearly outperforms the model trained on CEFR-SP underscoring the importance of data diversity in fine-tuning LMs for readability assessment.


\subsection{Zero-shot Cross Lingual Transfer}
\label{app:cross-lingual-transfer}

The zero-shot cross lingual results for several multilingual models are shown in Table~\ref{tab:cross-ling-all-models}. Similar to what is observed in \S\ref{sec:analyses}, fine-tuning on {\sc ReadMe++} leads to significantly better cross-lingual transfer to 6 different target languages compared to fine-tuning on previous datasets. The improvement and trend is consistent across various models. We provide in Table~\ref{tab:cross-lingual-domain} per-domain correlation results of XLMR$_{L}$ when transferring to Arabic, French, Hindi, and Russian, where we see superiority across domains by the model fine-tuned on  {\sc ReadMe++} compared with fine-tuning on the single-domain Wikipedia-based CEFR-SP.

\begin{table}[h!]
\centering
\begin{adjustbox}{width=\linewidth}
\renewcommand{\arraystretch}{0.9}
\begin{tabular}{@{}lrrrrrr@{}}
\toprule
\multicolumn{1}{c}{\multirow{2}{*}{\textbf{Model}}} & \multicolumn{2}{c}{\textbf{ReadMe++}} & \multicolumn{2}{c}{\textbf{CEFR-SP}} & \multicolumn{2}{c}{\textbf{CompDS}} \\ \cmidrule(l){2-7} 
\multicolumn{1}{c}{} & \multicolumn{1}{c}{\textbf{F1}} & \multicolumn{1}{c}{\textbf{$\rho$}} & \multicolumn{1}{c}{\textbf{F1}} & \multicolumn{1}{c}{\textbf{$\rho$}} & \multicolumn{1}{c}{\textbf{F1}} & \multicolumn{1}{c}{\textbf{$\rho$}} \\ \midrule
\textbf{en} $\rightarrow$ \textbf{ar} & \multicolumn{1}{l}{} & \multicolumn{1}{l}{} & \multicolumn{1}{l}{} & \multicolumn{1}{l}{} & \multicolumn{1}{l}{} & \multicolumn{1}{l}{} \\
mBERT & \textbf{19.94} & \textbf{0.512} & 12.38 & 0.368 & 1.76 & 0.099 \\
XLM-R$_{B}$ & \textbf{32.63} & \textbf{0.645} & 9.61 & 0.068 & 7.21 & 0.120 \\
XLM-R$_{L}$ & \textbf{31.48} & \textbf{0.606} & 8.81 &  0.071 & 5.99 & 0.322 \\ \midrule
\textbf{en} $\rightarrow$ \textbf{hi} &  &  &  &  &  &  \\
mBERT & \textbf{15.13} & \textbf{0.521} & 8.72 & 0.375 & 6.45 & 0.171 \\
XLM-R$_{B}$ & \textbf{16.57} & \textbf{0.655} & 9.87 & 0.146 & 9.81 & 0.398 \\
XLM-R$_{L}$ & \textbf{23.87} & \textbf{0.702} & 13.15 & 0.267 & 10.38 & 0.381 \\ \midrule
\textbf{en} $\rightarrow$ \textbf{fr} &  &  &  &  &  &  \\
mBERT & \textbf{30.63}   &  \textbf{0.751} & 10.87  & 0.490  & 8.02  & 0.341  \\
XLM-R$_{B}$ & \textbf{33.96}  & \textbf{0.746}  & 10.37  &  0.091 & 8.97 & 0.399 \\
XLM-R$_{L}$ & \textbf{30.29}  & \textbf{0.768}  &  11.06 & -0.026  & 5.92 &  0.335 \\ \midrule
\textbf{en} $\rightarrow$ \textbf{ru} &  &  &  &  &  &  \\
mBERT & \textbf{16.25}   & \textbf{0.610}  & 9.11   & 0.479   & 10.9   & 0.396 \\
XLM-R$_{B}$ & \textbf{21.27}  & \textbf{0.671}  &  13.16 & 0.253  & 12.64   &  0.404 \\
XLM-R$_{L}$ & \textbf{24.60}  & \textbf{0.760}  & 15.69  &  0.173 & 10.33  & 0.412 \\ \midrule
\textbf{en} $\rightarrow$ \textbf{it} &  &  &  &  &  &  \\
mBERT & \textbf{12.79} & \textbf{0.270} & 7.91 & 0.248 & 10.37 & 0.119 \\
XLM-R$_{B}$ & \textbf{14.38} & \textbf{0.295} & 9.66 & 0.029 & 12.00 & 0.137 \\
XLM-R$_{L}$ & \textbf{14.68} & \textbf{0.239} & 9.88 & -0.043 & 10.06 & 0.099 \\ \midrule
\textbf{en} $\rightarrow$ \textbf{de} &  &  &  &  &  &  \\
mBERT & \textbf{15.98} & \textbf{0.672} & 12.51 & 0.595 & 6.88 & 0.347 \\
XLM-R$_{B}$ & \textbf{27.13} & \textbf{0.702} & 14.02 & 0.196 & 8.68 & 0.529 \\
XLM-R$_{L}$ & \textbf{22.19} & \textbf{0.701} & 10.00 & -0.092 & 11.84 & 0.408 \\ \bottomrule
\end{tabular}
\end{adjustbox}
\caption{Zero-shot cross-lingual transfer performance. Models fine-tuned on English data (en) of {\sc ReadMe++} significantly outperform models fine-tuned with CEFR-SP \cite{cefr} or CompDS \cite{brunato} for Arabic (ar),  Hindi (hi), Italian (it), and German (de).}
\label{tab:cross-ling-all-models}
\end{table}

\begin{table*}
\centering
\begin{adjustbox}{width=\linewidth}
\begin{tabular}{@{}lcccccccc@{}}
\toprule
\multirow{2}{*}{\textbf{Domain}} & \multicolumn{2}{c}{\textbf{en} $\rightarrow$ \textbf{ar}} & \multicolumn{2}{c}{\textbf{en} $\rightarrow$ \textbf{fr}} & \multicolumn{2}{c}{\textbf{en} $\rightarrow$ \textbf{hi}} & \multicolumn{2}{c}{\textbf{en} $\rightarrow$ \textbf{ru}} \\ \cmidrule(l){2-9} 
 & \textbf{ReadMe++} & \textbf{CEFR-SP} & \textbf{ReadMe++} & \textbf{CEFR-SP} & \textbf{ReadMe++} & \textbf{CEFR-SP} & \textbf{ReadMe++} & \textbf{CEFR-SP} \\ \midrule
Captions & \textbf{0.545} & 0.165 & \textbf{0.551} & 0.179 & \textbf{0.336} & 0.028 & \textbf{0.644} & 0.202 \\
Dialogue & 0.126 & \textbf{0.269} & \textbf{0.635} & -0.387 & \textbf{0.438} & 0.122 & \textbf{0.150} & -0.220 \\
Dictionaries & -0.274 & \textbf{0.000} & --- & --- & --- & --- & --- & --- \\
Entertainment & \textbf{0.374} & 0.107 & 0.000 & 0.000 & \textbf{0.657} & 0.099 & \textbf{0.397} & 0.288 \\
Finance & --- & --- & \textbf{0.784} & -0.013 & --- & --- & \textbf{0.352} & -0.084 \\
Forums & \textbf{0.440} & 0.161 & \textbf{0.564} & 0.000 & \textbf{0.603} & 0.281 & \textbf{0.737} & -0.109 \\
Guides & \textbf{0.534} & 0.024 & \textbf{0.388} & -0.030 & \textbf{0.362} & 0.041 & \textbf{0.438} & 0.011 \\
Legal & \textbf{0.277} & -0.093 & \textbf{0.557} & -0.190 & \textbf{0.362} & 0.261 & \textbf{0.782} & -0.220 \\
Letters & --- & --- & \textbf{0.794} & 0.000 & --- & --- & \textbf{0.892} & 0.214 \\
Literature & \textbf{0.692} & 0.081 & \textbf{0.709} & -0.368 & \textbf{0.561} & 0.168 & \textbf{0.498} & 0.059 \\
News & \textbf{0.447} & 0.000 & --- & --- & --- & --- & --- & --- \\
Poetry & 0.000 & 0.000 & \textbf{0.339} & -0.068 & \textbf{0.202} & -0.347 & \textbf{0.779} & 0.112 \\
Policies & \textbf{0.835} & 0.009 & \textbf{0.727} & -0.070 & \textbf{0.551} & -0.427 & \textbf{0.703} & 0.144 \\
Research & \textbf{0.562} & -0.021 & \textbf{0.564} & 0.154 & \textbf{0.501} & -0.112 & \textbf{0.647} & 0.262 \\
Social Media & \textbf{0.620} & 0.313 & \textbf{0.489} & -0.677 & \textbf{0.341} & 0.036 & \textbf{0.452} & -0.106 \\
Speech & \textbf{0.337} & -0.147 & \textbf{0.618} & 0.291 & \textbf{0.668} & 0.200 & \textbf{0.583} & 0.118 \\
Statements & \textbf{0.374} & -0.019 & \textbf{0.592} & -0.193 & \textbf{0.331} & -0.013 & \textbf{0.602} & -0.130 \\
Textbooks & \textbf{0.600} & 0.569 & --- & --- & \textbf{0.427} & -0.201 & --- & --- \\
User Reviews & \textbf{0.570} & 0.240 & --- & --- & \textbf{0.375} & -0.018 & \textbf{0.000} & -0.196 \\
Wikipedia & \textbf{0.644} & 0.111 & \textbf{0.625} & 0.097 & \textbf{0.630} & 0.110 & \textbf{0.715} & 0.109 \\ \bottomrule
\end{tabular}
\end{adjustbox}
\caption{Pearson Correlation per domain when performing cross lingual transfer to Arabic, French, Hindi, and Russian using XLMR$_{L}$ fine-tuned with {\sc ReadMe++} (en) vs CEFR-SP-WikiAuto \cite{cefr}.}
\label{tab:cross-lingual-domain}
\end{table*}

\subsection{Effect of Context}
\label{app:context-effect}

\begin{figure}[h!]
    \centering
    \includegraphics[width=\linewidth]{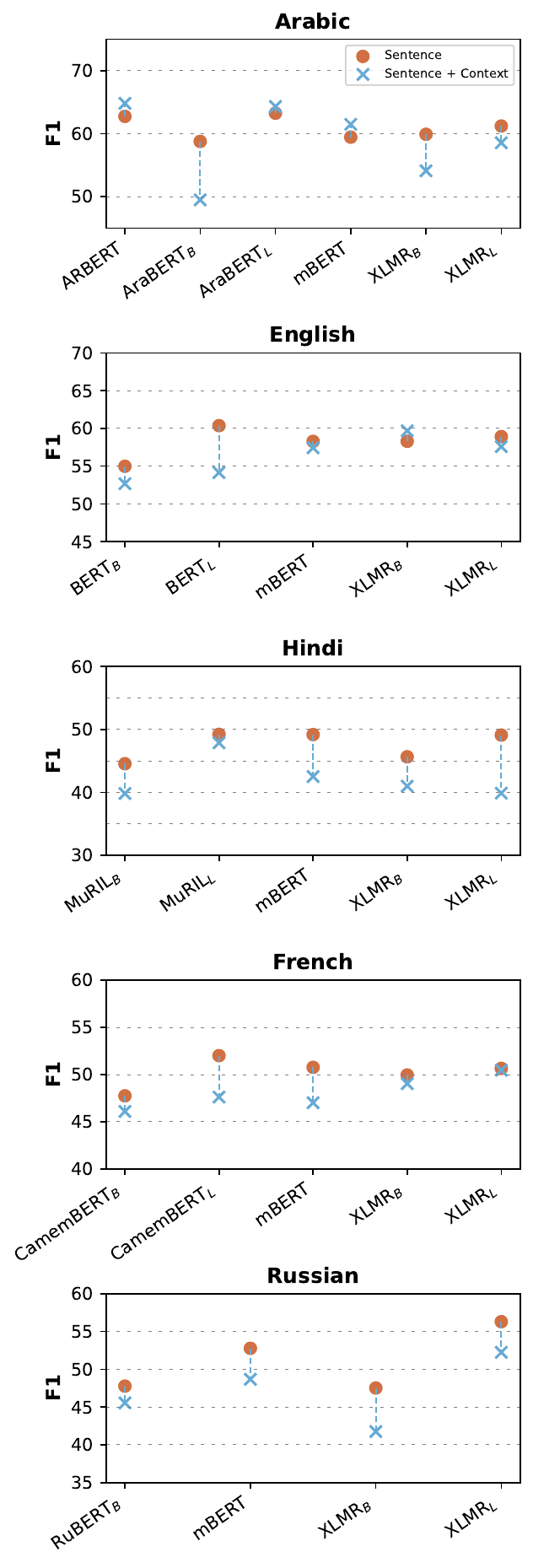}
    \caption{Effect of providing context during fine-tuning.}
    \label{fig:context-results}
\end{figure}

We study the effect of providing models with context during training, which consists of up to three sentences that precede a sentence lying within a paragraph, on performance in the supervised setting. We prepend the context to the input sentence when available and separate them with a \texttt{[SEP]} token. Figure~\ref{fig:context-results} shows the results with and without the addition of context when available. Overall, we find that pre-pending context information during fine-tuning decreased model performance in the majority of cases, or had little to no effect.

\section{Annotation Interface}
\label{appendix:interface}
\begin{figure*}[h!]
    \centering
    \includegraphics[width=\linewidth]{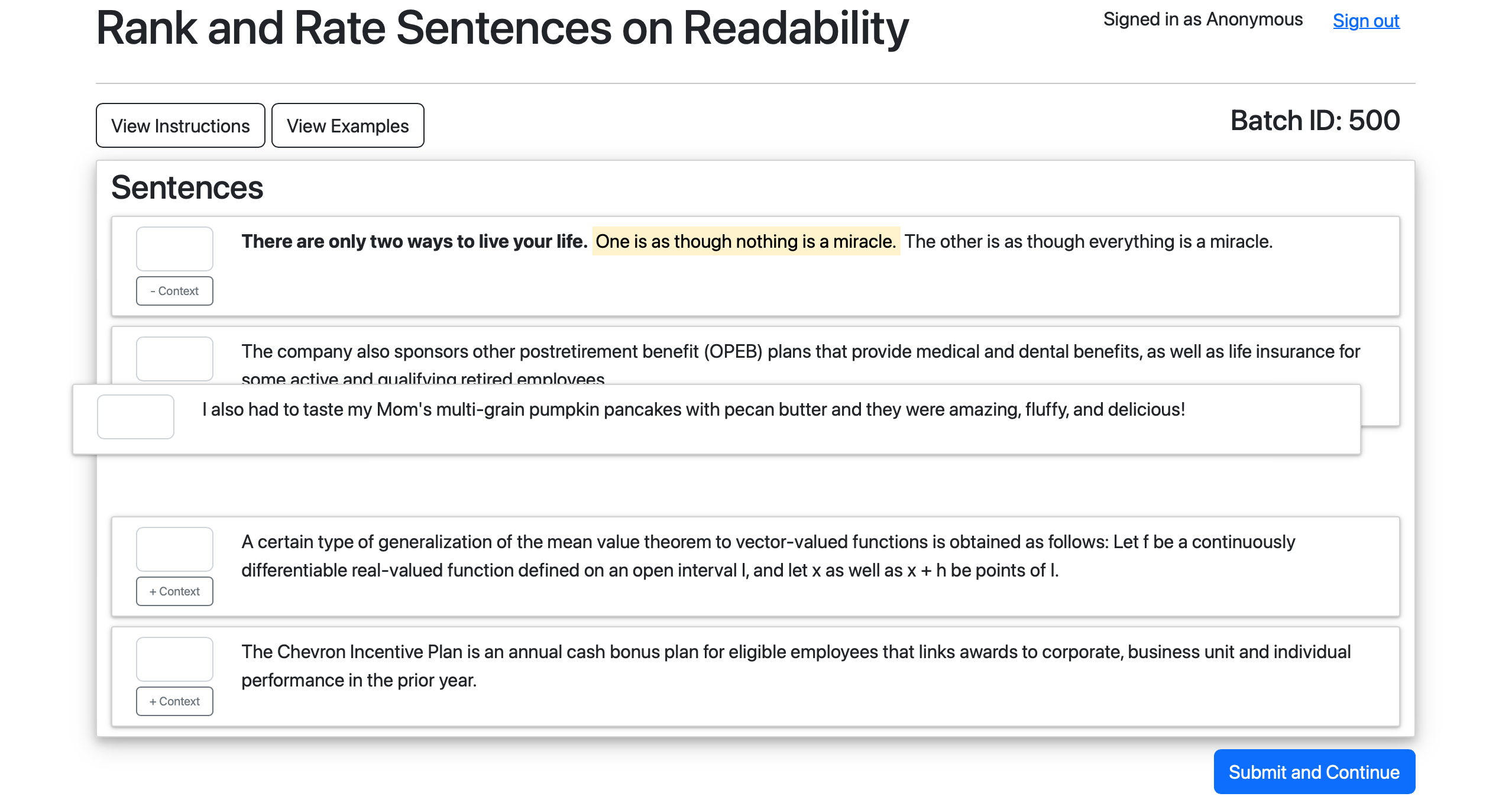}
    \caption{Screenshot of the developed annotation interface for rating English readability sentences. Annotators first rank sentences according to their readability level by simply dragging the box as shown in the figure. An optional Context button if available to show the context of a sentence if available.}
    \label{fig:english-rank}
\end{figure*}

\begin{figure*}[h!]
    \centering
    \includegraphics[width=\linewidth]{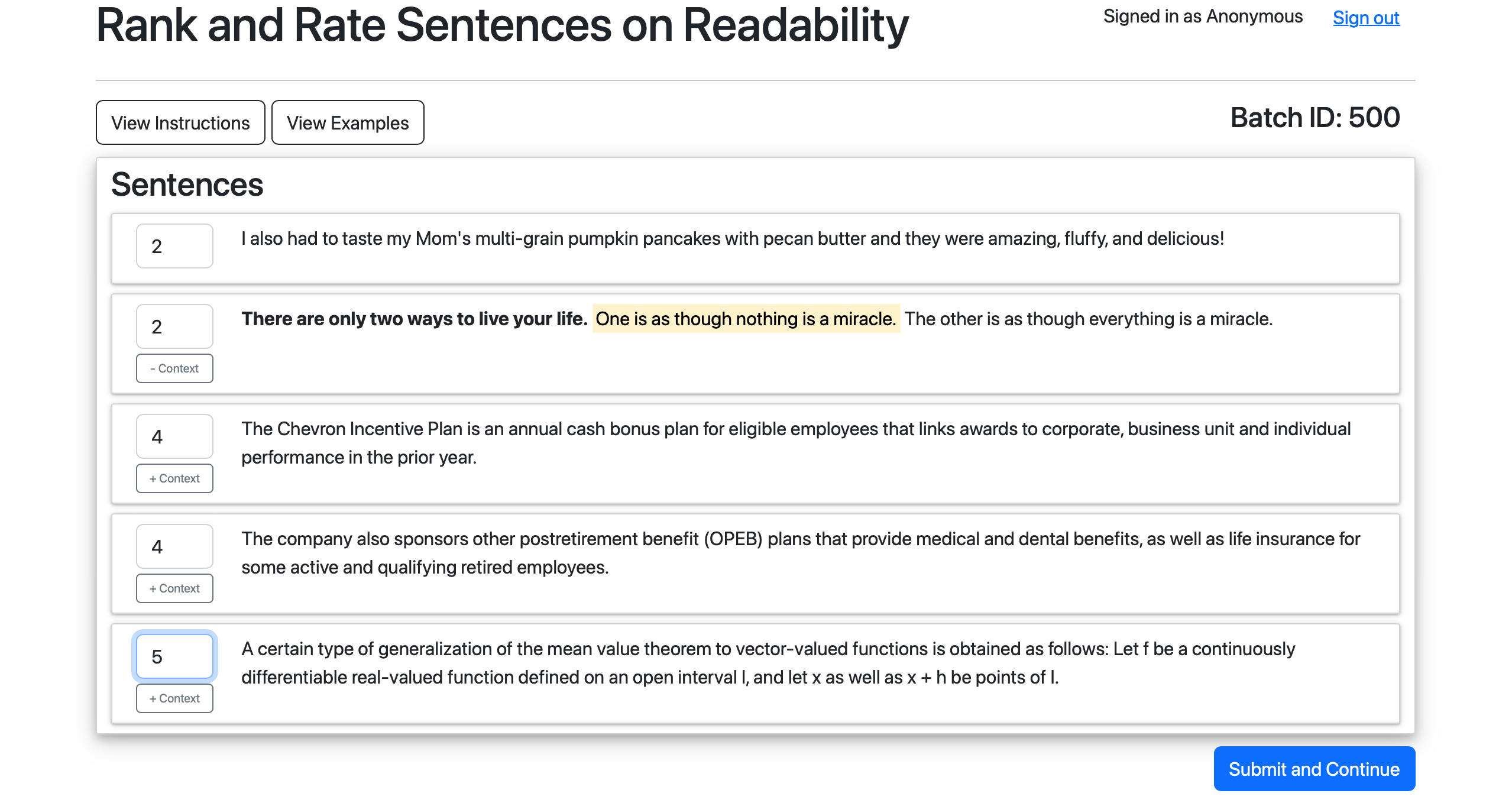}
    \caption{After ranking, annotators then assign a score for each sentence on a scale of 1 to 6 that corresponds to the CEFR levels. When done, annotators submit their scores and proceed to another batch of 5 sentences.}
    \label{fig:english-rate}
\end{figure*}

\begin{figure*}[h!]
    \centering
    \includegraphics[width=\linewidth]{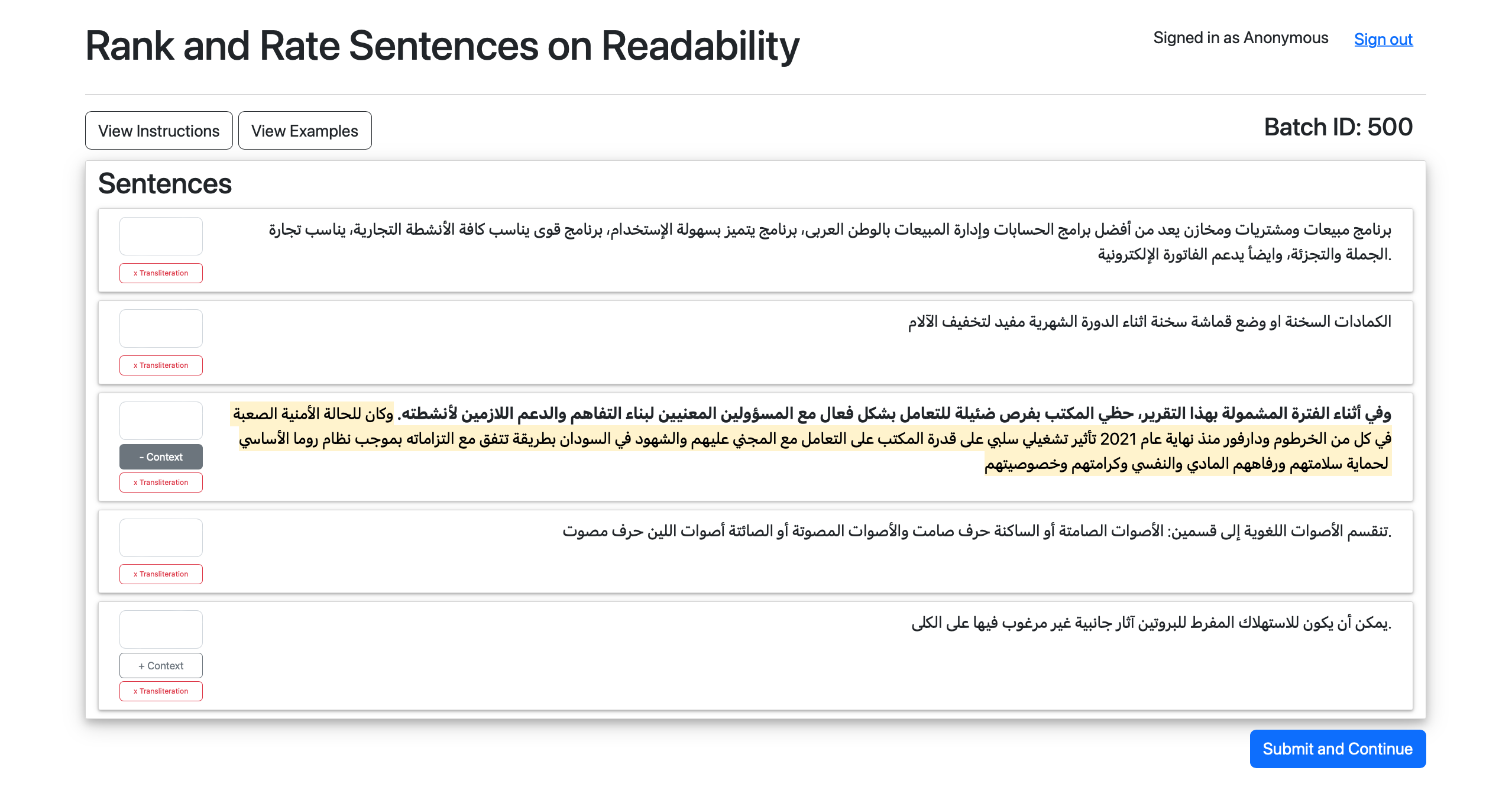}
    \caption{Screenshot of the developed annotation interface for Arabic sentences. An additional button to mark whether a sentence contains transliterations is provided.}
    \label{fig:arabic-interface}
\end{figure*}

\begin{figure*}[h!]
    \centering
    \includegraphics[width=\linewidth]{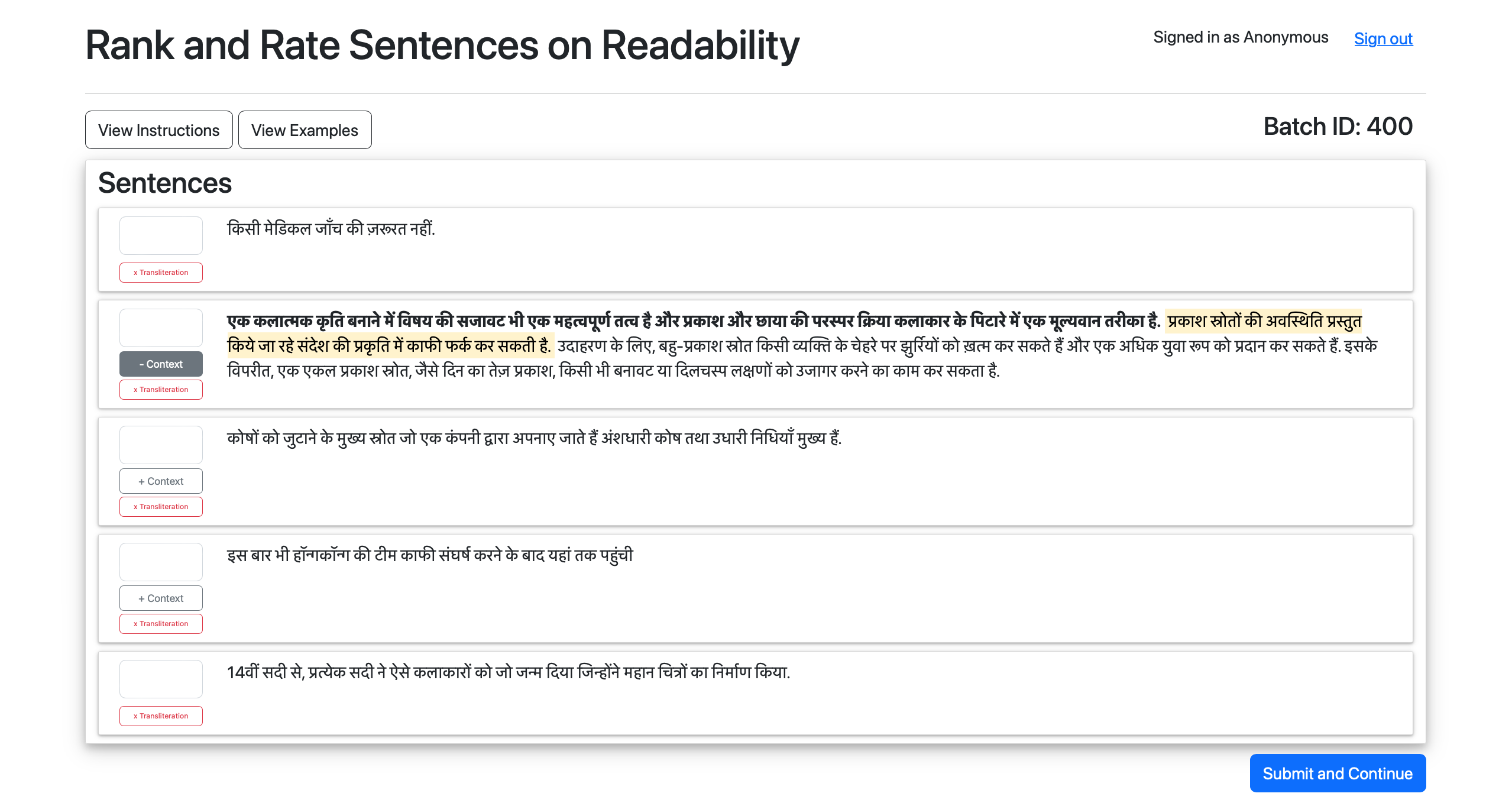}
    \caption{Screenshot of the developed annotation interface for Hindi sentences. An additional button to mark whether a sentence contains transliterations is provided.}
    \label{fig:hindi-interface}
\end{figure*}

Figures \ref{fig:english-rank} and \ref{fig:english-rate} show screenshots of our developed annotation interface for English sentences, where annotators perform a rank-and-rate approach to assign readability scores to 5 sentences in each batch. Annotators are asked to first rank sentences which they can do by simply dragging them. They are then asked to choose a rating for each sentence from a drop-down list. For each sentence, we provide the option to show its context, which shows the sentence in the paragraph to which it belongs. Figures~\ref{fig:arabic-interface} and \ref{fig:hindi-interface} show screenshots of the interface for Arabic and Hindi respectively. An additional button to mark transliterations is added.

\section{License and Use Terms}
\label{appendix:license}
We provide in Tables~\ref{tab:license-1}, \ref{tab:license-2}, and \ref{tab:license-3} the license  or usage term for each data source used in the creation of the corpus as follows:

\begin{itemize}
    \item License: exact license under which data is available (CC BY 4.0 or other).
    
    \item Public Domain: data available in the public domain.
    
    \item Personal/Non-Commercial: source grants usage permission of data for personal/non-commercial purposes.
    
    \item (---): denotes that data needs to be requested from authors.
\end{itemize}

\begin{table*}[]
\centering
\begin{adjustbox}{width=1\textwidth}
\begin{tabular}{@{}l@{}}
\toprule
{\sc Literature} - Novels \\
 \\
\begin{tabular}[c]{@{}l@{}}Over the river men were at work with spades and sieves on the sandy foreshore, and on the river was a boat, also diligently employed\\  for some mysterious end. An electric tram came rushing underneath the window. \sethlcolor{lightgreen}\hl{No one was inside it, except one tourist;} \\ \sethlcolor{lightgreen}\hl{but its platforms were overflowing with Italians, who preferred to stand. Children tried to hang on behind, and the  conductor,}\\ \sethlcolor{lightgreen}\hl{with no malice, spat in their faces to make them let go. Then soldiers appeared--good-looking, undersized men--wearing}\\ \sethlcolor{lightgreen}\hl{each a knapsack covered with mangy fur, and a great-coat which had been cut for some larger soldier.} \sethlcolor{lightblue}\hl{Beside them walked}\\ \sethlcolor{lightblue}\hl{officers, looking foolish and fierce, and before them went little boys, turning somersaults in time with the band.} The tramcar\\ became entangled in their ranks, and moved on painfully, like a caterpillar in a swarm of ants. One of the little boys fell down,\\ and some white bullocks came out of an archway. Indeed, if it had not been for the good advice of an old man who was selling \\ button-hooks, the road might never have got clear.\end{tabular} \\
 \\ \midrule
{\sc Medical} - Clinical Reports \\
 \\
\begin{tabular}[c]{@{}l@{}}\sethlcolor{lightblue}\hl{The patient underwent a flex sigmoidoscopy on Friday , 11-02 , which showed old blood in the rectal vault but no active source of bleeding.}\\ Given this , it was advised that the patient have a colonoscopy to rule out further bleeding\end{tabular} \\
 \\ \midrule
{\sc Textbooks} - Engineering \\
 \\
\begin{tabular}[c]{@{}l@{}}\sethlcolor{lightgreen}\hl{The script might email information about the target user to the attacker, or might attempt to exploit a browser vulnerability on the target}\\ \sethlcolor{lightgreen}\hl{system in order to take it over completely.} \sethlcolor{lightblue}\hl{The script and its enclosing tags will not appear in what the victim actually sees on the screen.}\end{tabular} \\
 \\ \midrule
{\sc Forums} - StackOverflow \\
 \\
What's the best way to convert a string to an enumeration value in C\# ? \\
 \\ \midrule
{\sc User Reviews} - Product \\
 \\
\begin{tabular}[c]{@{}l@{}}\sethlcolor{lightblue}\hl{First of all the package was shoved into my mail box and was basically crushed when I pulled it out.} In addition there are deep marks and scrapes \\ that show the wallet was used or pre-owned before getting to me..\end{tabular} \\
 \\ \midrule
{\sc Statements} - Quotes \\
 \\
I may not have gone where I intended to go, but I think I have ended up where I needed to be. \\
 \\ \midrule
{\sc Wikipedia} - Philosophy \\
 \\
\begin{tabular}[c]{@{}l@{}}Monarchies are associated with hereditary reign, in which monarchs reign for life and the responsibilities and power of the position pass to their child\\  or another member of their family when they die.\end{tabular} \\
 \\ \bottomrule
\end{tabular}
\end{adjustbox}
\caption{English Examples from several domains of {\sc ReadMe++}. The sentence annotated for readability is highlighted in  blue within the paragraph it belongs to, if applicable. Up to three preceding sentences of context to the sentence are highlighted in green if applicable.}
\label{tab:examples-en}
\end{table*}

\begin{table*}[h]
\setcode{utf8}
\centering
\begin{adjustbox}{width=\linewidth}
\begin{tabular}{@{}l@{}}
\toprule
{\sc Literature} - History \\
 \\
\multicolumn{1}{r}{\small{\<بل لقد كانت بدر بمثابة العَلَم الخفَّاق الذي يُرفرِف على ممتلكات الإسلام في قابل السنين والأعوام، كانت بدايةَ فتح خير دِين سمت مبادؤه، وتلألأت أضواؤه>}} \\
\multicolumn{1}{r}{\small{\<حتى بلغت جبال الألب والبيرنيه غربًا، والصين واليابان شرقًا، وصار معتنقوه خمسمائة مليون من النفوس بعد أن كانوا نفرًا قليلًا، محمدًا وصحبه الأكرمين الأولين>}} \\
 \\
\begin{tabular}[c]{@{}l@{}}\textit{Translation:} Rather, Badr was like a fluttering flag that flutters over the possessions of Islam in the face of years and years. It was the beginning\\  of the conquest of the best religion whose principles were elevated, and its lights sparkled. It reached the Alps and the Pyrenees in the west, and China\\  and Japan in the east, and its adherents became five hundred million souls after they were a small number; Muhammad and his first noble companions.\end{tabular} \\
\multicolumn{1}{r}{} \\ \midrule
{\sc News Articles} - Sports \\
\multicolumn{1}{r}{\small{\<يستضيف ملعب كامب نو اليوم السبت انطلاقا من الساعة مساء نهائي كأس الملك بين برشلونة وأتلتيك بلباو فيما يلي التشكيلة المتوقعة بحسب صحيفة موندو ديبرتيفو>}} \\
\begin{tabular}[c]{@{}l@{}}\textit{Translation:} Today, Saturday, the Camp Nou stadium will host the King’s Cup final between Barcelona and Athletic Bilbao. The following is the\\  expected line-up,  according to the Mundo Deportivo newspaper.\end{tabular} \\
 \\ \midrule
{\sc Policies} - Contracts \\
 \\
\multicolumn{1}{r}{\small{\<جميع المصاريف والأتعاب الناشئة عن مماطلة أيّ من الطرفين في سداد الأقساط أو سداد مصاريف الصيانة، أو إزالة الضرر الناشئ بسببه>}} \\
\multicolumn{1}{r}{\small{\<تعتبر جزءًا من التزاماته الأصلية، ويتعهد الطرف المماطل بدفعها>}} \\
 \\
\begin{tabular}[c]{@{}l@{}}\textit{Translation:} All expenses and fees arising from the delay of either party in paying the installments or paying the maintenance expenses,\\  or removing the damage arising because of it, are considered part of their original obligations, and the party that caused the delay\\  undertakes to pay them\end{tabular} \\
 \\ \midrule
{\sc Guides} - Online Tutorials \\
 \\
\multicolumn{1}{r}{\small{\<يجب أن تضع الطائر بعيدًا عن الأطفال الصغار أو أي حيوانات أخرى قد تهاجمه أو تصيبه بإصابة أخرى دون قصد>}} \\
 \\
\textit{Translation:} You should keep the bird away from small children or other animals that might attack or otherwise inadvertently injure it \\
 \\ \midrule
{\sc Dictionaries} \\
 \\
\multicolumn{1}{r}{\small{\<ألا إن شر الروايا روايا الكذب>}} \\
\textit{Translation:} Verily, the most evil of stories are false stories \\
 \\ \midrule
{\sc Statements} - Quotes \\
 \\
\multicolumn{1}{r}{\small{\<العاقل لا يستقبل النعمة ببطر ولا يودعها بجزع>}} \\
 \\
\textit{Translation:} The wise person does not welcome a blessing with arrogance, nor does he become impatient when he loses it \\
 \\ \midrule
{\sc Poetry} \\
\multicolumn{1}{r}{} \\
\multicolumn{1}{r}{\small{\<أَرِقتُ لَهُ وَالبَرقُ دونَ طَمِيَّةٍ>}} \\
 \\ \bottomrule
\end{tabular}
\end{adjustbox}
\caption{Arabic sentence examples from {\sc ReadMe++}. Note that a sentence in Arabic could be translated into multiple sentences in English.}
\label{tab:examples-ar}
\end{table*}

\newpage
\clearpage

\newpage

\begin{table*}[h!]
\centering
{\tiny\renewcommand{\arraystretch}{.95}
\resizebox{!}{.25\paperheight}{%
\begin{tabular}{rccccllrccccl}
\cline{1-6} \cline{8-13}
\multicolumn{1}{l}{\textbf{Domain}} & \multicolumn{5}{c}{\textbf{\# Sentences}} &  & \multicolumn{1}{l}{\textbf{Domain}} & \multicolumn{5}{c}{\textbf{\# Sentences}} \\ \cline{1-6} \cline{8-13} 
\textbf{Sub-Domain} & \textbf{ar} & \textbf{en} & \multicolumn{1}{l}{\textbf{fr}} & \textbf{hi} & \textbf{ru} &  & \textbf{Sub-Domain} & \textbf{ar} & \textbf{en} & \textbf{fr} & \textbf{hi} & \textbf{ru} \\ \cline{1-6} \cline{8-13} 
\multicolumn{6}{l}{{\sc Wikipedia}} &  & \multicolumn{6}{l}{{\sc Forums}} \\
History & 50 & 50 & 50 & 22 & 50 &  & Reddit & 39 & 50 & \multicolumn{1}{l}{50} & 49 & 50 \\
Geography & 50 & 50 & 50 & 31 & 50 &  & QA Websites & 28 & 48 & \multicolumn{1}{l}{50} & 47 & 50 \\
Philosophy & 49 & 47 & 50 & 34 & 50 &  & StackOverflow & --- & 50 & --- & --- & \multicolumn{1}{c}{---} \\ \cline{8-13} 
Technology & 43 & 50 & 50 & 19 & 50 &  & \multicolumn{6}{l}{{\sc Social Media}} \\
Mathematics & 43 & 50 & 32 & 23 & 50 &  & Twitter & 41 & 47 & \multicolumn{1}{l}{50} & 44 & 49 \\ \cline{8-13} 
Art \& Culture & 49 & 50 & 50 & 35 & 50 &  & \multicolumn{6}{l}{{\sc Policies}} \\
Social Sciences & 48 & 50 & 50 & 41 & 50 &  & Contracts & 27 & 34 & 45 & --- & \multicolumn{1}{c}{41} \\
Natural Sciences & 49 & 49 & 50 & 38 & 50 &  & Olympic Rules & 40 & 50 & 50 & --- & \multicolumn{1}{c}{50} \\
Health \& Fitness & 49 & 49 & 50 & 40 & 50 &  & Code of Conduct & --- & 50 & --- & 50 & \multicolumn{1}{c}{---} \\ \cline{1-6} \cline{8-13} 
\multicolumn{6}{l}{{\sc News Articles}} &  & \multicolumn{6}{l}{{\sc Guides}} \\
Sports & 46 & 46 & --- & --- & \multicolumn{1}{c}{---} &  & User Manuals & 50 & 46 & 50 & 28 & 50 \\
Politics & 13 & 44 & --- & --- & \multicolumn{1}{c}{---} &  & Online Tutorials & 51 & 47 & 50 & 44 & 50 \\
Culture & 50 & 50 & --- & --- & \multicolumn{1}{c}{---} &  & Cooking Recipes & 40 & 48 & 50 & 47 & 50 \\
Economy & 41 & 50 & --- & --- & \multicolumn{1}{c}{---} &  & Code Documentation & --- & 49 & --- & --- & \multicolumn{1}{c}{---} \\ \cline{8-13} 
Technology & 36 & 50 & --- & --- & \multicolumn{1}{c}{---} &  & \multicolumn{6}{l}{{\sc Captions}} \\ \cline{1-6}
\multicolumn{6}{l}{{\sc Research}} &  & Images & 50 & 50 & 47 & 48 & 44 \\
Law & 36 & 19 & --- & 13 & 50 &  & Videos & --- & 50 & 50 & 50 & \multicolumn{1}{c}{---} \\
Politics & 19 & 22 & --- & 19 & 50 &  & Movies & 27 & 41 & 50 & 46 & \multicolumn{1}{c}{---} \\
Medical & --- & 30 & 31 & --- & 50 &  & YouTube & --- & 42 & --- & --- & \multicolumn{1}{c}{---} \\ \cline{8-13} 
Literature & --- & 39 & --- & 28 & \multicolumn{1}{c}{---} &  & \multicolumn{6}{l}{{\sc Medical Text}} \\
Economics & 26 & 46 & --- & 31 & 50 &  & Clinical Reports & --- & 39 & --- & --- & \multicolumn{1}{c}{---} \\ \cline{8-13} 
Science \& Engineering & --- & 30 & 47 & --- & 50 &  & \multicolumn{6}{l}{{\sc Entertainment}} \\ \cline{1-6}
\multicolumn{6}{l}{{\sc Literature}} &  & Jokes & 50 & 50 & --- & 46 & 49 \\ \cline{8-13} 
Novels & 50 & 50 & 50 & 48 & 50 &  & \multicolumn{6}{l}{{\sc Speech}} \\
History & 40 & 45 & 50 & 47 & \multicolumn{1}{c}{---} &  & Ted Talks & 49 & 43 & \multicolumn{1}{l}{50} & 48 & 50 \\
Biographies & 26 & 47 & --- & 46 & \multicolumn{1}{c}{---} &  & Public Speech & 35 & 47 & --- & 45 & 30 \\ \cline{8-13} 
Children’s Books & 50 & 49 & 50 & 44 & \multicolumn{1}{c}{---} &  & \multicolumn{6}{l}{{\sc Statements}} \\ \cline{1-6}
\multicolumn{6}{l}{{\sc Textbooks}} &  & Rumours & 20 & 40 & --- & 39 & \multicolumn{1}{c}{---} \\
Business & 35 & 50 & \multicolumn{1}{l}{---} & 47 & --- &  & Quotes & 50 & 50 & 50 & 49 & 50 \\ \cline{8-13} 
Psychology & --- & 50 & \multicolumn{1}{l}{---} & 47 & --- &  & \multicolumn{6}{l}{{\sc Dialogue}} \\
Agriculture & --- & 50 & \multicolumn{1}{l}{---} & --- & --- &  & Open-domain & 39 & 44 & \multicolumn{1}{l}{50} & 39 & 49 \\
Engineering & --- & 50 & \multicolumn{1}{l}{---} & --- & --- &  & Negotiation & --- & 45 & --- & --- & \multicolumn{1}{c}{---} \\ \cline{1-6}
\multicolumn{6}{l}{{\sc User Reviews}} &  & Task-oriented & 39 & 50 & \multicolumn{1}{l}{50} & 50 & \multicolumn{1}{c}{---} \\ \cline{8-13} 
Products & 50 & 40 & --- & 33 & 49 &  & \multicolumn{6}{l}{{\sc Legal}} \\
Books & 50 & 47 & --- & --- & \multicolumn{1}{c}{---} &  & Constitutions & 43 & 30 & \multicolumn{1}{l}{50} & 34 & 50 \\
Movies & --- & 50 & --- & 43 & \multicolumn{1}{c}{---} &  & Judicial Rulings & --- & 21 & --- & 35 & 47 \\
Hotels & 50 & 48 & --- & --- & \multicolumn{1}{c}{---} &  & UN Parliament & 39 & 43 & \multicolumn{1}{l}{50} & --- & 50 \\ \cline{8-13} 
Restaurants & 50 & 47 & --- & --- & \multicolumn{1}{c}{---} &  & \multicolumn{1}{l}{{\sc Finance}} & --- & 50 & \multicolumn{1}{l}{50} & --- & 50 \\ \cline{1-6} \cline{8-13} 
\multicolumn{1}{l}{{\sc Dictionaries}} & 40 & 40 & --- & --- & \multicolumn{1}{c}{---} &  & \multicolumn{1}{l}{{\sc Poetry}} & 46 & 50 & \multicolumn{1}{l}{50} & 49 & 50 \\ \cline{1-6} \cline{8-13} 
 &  &  & \multicolumn{1}{l}{} &  &  &  & \multicolumn{1}{l}{{\sc Letters}} & --- & 22 & \multicolumn{1}{l}{50} & --- & 50 \\ \cline{8-13} 
\end{tabular}
}}
\caption{Dataset Statistics. (---)  denotes that no public resource was found in the particular language.}
\label{tab:dataset-statistics}
\end{table*}

\begin{table*}[h!]
\centering
{\tiny\renewcommand{\arraystretch}{.9}
\resizebox{!}{.35\paperheight}{%
\begin{tabular}{@{}rlll@{}}
\toprule
\multicolumn{1}{l}{\textbf{Domain}} & \multicolumn{3}{c}{\textbf{Source}} \\ \midrule
\textbf{Sub-Domain} & \textbf{ar} & \textbf{en} & \textbf{hi} \\ \midrule
\multicolumn{1}{l}{{\sc Wikipedia}} & \cellcolor[HTML]{EFEFEF}wikipedia.com & \cellcolor[HTML]{EFEFEF}wikipedia.com & \cellcolor[HTML]{EFEFEF}wikipedia.com \\ \midrule
\multicolumn{1}{l}{{\sc News Articles}} & \cite{arabic-news} & \cite{english-news} & --- \\ \midrule
\multicolumn{4}{l}{{\sc Research}} \\
Law & \cellcolor[HTML]{EFEFEF}spu.sharjah.ac.ae & \cellcolor[HTML]{EFEFEF}elgaronline.com & \cellcolor[HTML]{EFEFEF}library.bjp.org \\
Politics & jcopolicy.uobaghdad.edu.iq & tandfonline.com & journal.ijarms.org \\
Medical & \cellcolor[HTML]{EFEFEF}--- & \cellcolor[HTML]{EFEFEF}onlinelibrary.wiley.com & \cellcolor[HTML]{EFEFEF}--- \\
Literature & --- & jstor.org/journal/jmodelite & hindijournal.com \\
Economics & \cellcolor[HTML]{EFEFEF}asjp.cerist.dz/index.php/en & \cellcolor[HTML]{EFEFEF}aeaweb.org & \cellcolor[HTML]{EFEFEF}journal.ijarms.org \\
Science \& Engineering & --- & arxiv.org & --- \\ \midrule
\multicolumn{1}{l}{{\sc Literature}} & \cellcolor[HTML]{EFEFEF}hindawi.org/books/ & \cellcolor[HTML]{EFEFEF}gutenberg.org & \cellcolor[HTML]{EFEFEF}Public Domain Books \\ \midrule
\multicolumn{1}{l}{{\sc Textbooks}} & hindawi.org/books/ & open.umn.edu & ncert.nic.in \\ \midrule
\multicolumn{4}{l}{{\sc Legal}} \\
Constitutions & \cellcolor[HTML]{EFEFEF}presidency.gov.lb & \cellcolor[HTML]{EFEFEF}constitutioncenter.org & \cellcolor[HTML]{EFEFEF}legislative.gov.in \\
Judicial Rulings & --- & law.cornell.edu/supremecourt & HLDC \cite{judicial-hi} \\
UN Parliament & \multicolumn{2}{c}{\cellcolor[HTML]{EFEFEF}United Nations Parallel Corpus \cite{un-parallel-corpus}} & \cellcolor[HTML]{EFEFEF}--- \\ \midrule
\multicolumn{1}{l}{{\sc User Reviews}} &  &  &  \\
Products & \cellcolor[HTML]{EFEFEF}\cite{reviews-ar} & \cellcolor[HTML]{EFEFEF}MARC \cite{product-review-en} & \cellcolor[HTML]{EFEFEF}\cite{review-product-hi} \\
Books & LABR \cite{book-review-ar} & \cite{good-reads} & --- \\
Movies & \cellcolor[HTML]{EFEFEF}--- & \cellcolor[HTML]{EFEFEF}JMURv1 \cite{movie-review-en} & \cellcolor[HTML]{EFEFEF}\cite{hindi-movie-reviews} \\
Hotels & \cite{reviews-ar} & \cite{hotel-reviews-en} & --- \\
Restaurants & \cellcolor[HTML]{EFEFEF}\cite{reviews-ar} & \cellcolor[HTML]{EFEFEF}\cite{trip-advisor-dataset} & \cellcolor[HTML]{EFEFEF}--- \\ \midrule
\multicolumn{4}{l}{{\sc Dialogue}} \\
Open-domain & ArabicED \cite{open-domain-ar} & DailyDialog \cite{open-domain-en} & MDIA \cite{open-domain-hi} \\
Negotiation & \cellcolor[HTML]{EFEFEF}--- & \cellcolor[HTML]{EFEFEF}CraigslistBargain \cite{negotiation-en} & \cellcolor[HTML]{EFEFEF}--- \\
Task-oriented & xSID \cite{task-oriented-en} & xSID \cite{task-oriented-en} & HDRS \cite{task-oriented-hi} \\ \midrule
\multicolumn{1}{l}{{\sc Forums}} &  &  &  \\
Reddit & \multicolumn{3}{c}{\cellcolor[HTML]{EFEFEF}Reddit Dump} \\
QA Websites & CQA-MD \cite{qa-ar} & quora.com \cite{qa-en} & \cite{hindi-qa} \\
StackOverflow & \cellcolor[HTML]{EFEFEF}--- & \cellcolor[HTML]{EFEFEF}\cite{stackoverflow} & \cellcolor[HTML]{EFEFEF}--- \\ \midrule
\multicolumn{4}{l}{{\sc Social Media}} \\
Twitter & \multicolumn{3}{c}{Stanceosaurus \cite{stanceosaurus}} \\ \midrule
\multicolumn{4}{l}{{\sc Policies}} \\
Contracts & \cellcolor[HTML]{EFEFEF}ejar.sa & \cellcolor[HTML]{EFEFEF}honeybook.com & \cellcolor[HTML]{EFEFEF}--- \\
Olympic Rules & \multicolumn{2}{c}{resources.specialolympics.org/translated-resources} & --- \\
Code of Conduct & \cellcolor[HTML]{EFEFEF}--- & \cellcolor[HTML]{EFEFEF}fatimafellowship.com & \cellcolor[HTML]{EFEFEF}lonza.com \\ \midrule
\multicolumn{4}{l}{{\sc Guides}} \\
User Manuals & \multicolumn{3}{c}{samsung.com/us/support/downloads} \\
Online Tutorials & \cellcolor[HTML]{EFEFEF}ar.wikihow.com & \cellcolor[HTML]{EFEFEF}wikihow.com & \cellcolor[HTML]{EFEFEF}hi.wikihow.com \\
Cooking Recipes & ar.wikibooks.org & en.wikibooks.org & --- \\
Code Documentation & \cellcolor[HTML]{EFEFEF}--- & \cellcolor[HTML]{EFEFEF}mathworks.com & \cellcolor[HTML]{EFEFEF}--- \\ \midrule
\multicolumn{4}{l}{{\sc Captions}} \\
Images & \cite{image-captions-ar} & Flikr30K \cite{image-captions-en} & \cite{image-captions-hi} \\
Videos & \cellcolor[HTML]{EFEFEF}--- & \cellcolor[HTML]{EFEFEF}Vatex \cite{video-captions-en} & \cellcolor[HTML]{EFEFEF}\cite{video-captions-hi} \\
Movies & \multicolumn{3}{c}{OpenSubtitles2016 \cite{open-subtitles}} \\
YouTube & \cellcolor[HTML]{EFEFEF}--- & \cellcolor[HTML]{EFEFEF}youtube.com & \cellcolor[HTML]{EFEFEF}--- \\ \midrule
\multicolumn{4}{l}{{\sc Medical Text}} \\
Clinical Reports & --- & i2b2/VA \cite{clinical-reports-en} & --- \\ \midrule
\multicolumn{1}{l}{{\sc Dictionaries}} & \cellcolor[HTML]{EFEFEF}almaany.com & \cellcolor[HTML]{EFEFEF}dictionary.com & \cellcolor[HTML]{EFEFEF}--- \\ \midrule
\multicolumn{4}{l}{{\sc Entertainment}} \\
Jokes & \cite{arabic-humour} & \cite{jokes-en} & 123hindijokes.com \\ \midrule
\multicolumn{1}{l}{{\sc Finance}} & \cellcolor[HTML]{EFEFEF}--- & \cellcolor[HTML]{EFEFEF}\cite{finance} & \cellcolor[HTML]{EFEFEF}--- \\ \midrule
\multicolumn{4}{l}{{\sc Speech}} \\
Ted Talks & ted.com/talks & ted.com/talks & ted.com/talks \\
Public Speech & \cellcolor[HTML]{EFEFEF}state.gov/translations/arabic & \cellcolor[HTML]{EFEFEF}whitehouse.gov & \cellcolor[HTML]{EFEFEF}--- \\ \midrule
\multicolumn{4}{l}{{\sc Statements}} \\
Rumours & \multicolumn{3}{c}{Stanceosaurus \cite{stanceosaurus}} \\
Quotes & \cellcolor[HTML]{EFEFEF}arabic-quotes.com & \cellcolor[HTML]{EFEFEF}goodreads.com/quotes & \cellcolor[HTML]{EFEFEF}storyshala.in \\ \midrule
\multicolumn{1}{l}{{\sc Poetry}} & aldiwan.net & poetryfoundation.org & hindionlinejankari.com \\ \midrule
\multicolumn{1}{l}{{\sc Letters}} & \cellcolor[HTML]{EFEFEF}--- & \cellcolor[HTML]{EFEFEF}oflosttime.com & \cellcolor[HTML]{EFEFEF}--- \\ \bottomrule
\end{tabular}
}}
\caption{Dataset Sources (1/2). (---)  denotes that no resource was found in the particular language.}
\label{tab:dataset-sources}
\end{table*}

\begin{table*}[]
\centering
\begin{tabular}{rll}
\hline
\multicolumn{1}{l}{\textbf{Domain}} & \multicolumn{2}{c}{\textbf{Source}} \\ \hline
\textbf{Sub-Domain} & \textbf{fr} & \textbf{ru} \\ \hline
\multicolumn{1}{l}{{\sc Wikipedia}} & \cellcolor[HTML]{EFEFEF}wikipedia.com & \cellcolor[HTML]{EFEFEF}wikipedia.com \\ \hline
\multicolumn{1}{l}{{\sc Research}} & hal.science & ruscorpora.ru \\
\multicolumn{1}{l}{{\sc Literature}} & \cellcolor[HTML]{EFEFEF}gutenberg.org & \cellcolor[HTML]{EFEFEF}gutenberg.org \\ \hline
\multicolumn{1}{l}{{\sc Legal}} &  &  \\
Constitutions & \cellcolor[HTML]{EFEFEF}legifrance.gouv.fr & \cellcolor[HTML]{EFEFEF}constitution.ru \\
Judicial Rulings & --- & supcourt.ru \\
UN Parliament & \multicolumn{2}{c}{\cellcolor[HTML]{EFEFEF}United Nations Parallel Corpus \cite{un-parallel-corpus}} \\ \hline
\multicolumn{1}{l}{{\sc User Reviews}} &  &  \\
Products & --- & RuReviews \cite{ru-product-review} \\ \hline
\multicolumn{1}{l}{{\sc Dialogue}} &  &  \\
Open-domain & \cellcolor[HTML]{EFEFEF}MDIA \cite{open-domain-hi} & \cellcolor[HTML]{EFEFEF}MDIA \cite{open-domain-hi} \\
Task-oriented & M-CID \cite{task-oriented-fr} & --- \\ \hline
\multicolumn{1}{l}{{\sc Forums}} &  &  \\
Reddit & \multicolumn{2}{c}{\cellcolor[HTML]{EFEFEF}Reddit Dump} \\
QA Websites & \cite{qa-fr} & \cite{qa-ru} \\ \hline
\multicolumn{1}{l}{{\sc Social Media}} &  &  \\
Twitter & \cellcolor[HTML]{EFEFEF}\cite{twitter-fr} & \cellcolor[HTML]{EFEFEF}RuSentiTweet \cite{twitter-ru} \\ \hline
\multicolumn{1}{l}{{\sc Policies}} &  &  \\
Contracts & \cellcolor[HTML]{EFEFEF}cesu.urssaf.fr & \cellcolor[HTML]{EFEFEF}blanker.ru \\
Olympic Rules & \multicolumn{2}{c}{resources.specialolympics.org/translated-resources} \\ \hline
\multicolumn{1}{l}{{\sc Guides}} &  &  \\
User Manuals & \cellcolor[HTML]{EFEFEF}samsung.com/us/support/downloads & \cellcolor[HTML]{EFEFEF}manuals.plus/ru \\
Online Tutorials & \multicolumn{2}{c}{wikihow.com} \\
Cooking Recipes & \multicolumn{2}{c}{\cellcolor[HTML]{EFEFEF}wikibooks.org} \\ \hline
\multicolumn{1}{l}{{\sc Captions}} &  &  \\
Images & \multicolumn{2}{c}{\cellcolor[HTML]{EFEFEF}\cite{image-captions-fr-ru}} \\
Videos & cite{video-captions-fr} & --- \\
Movies & \multicolumn{2}{c}{\cellcolor[HTML]{EFEFEF}OpenSubtitles2016 \cite{open-subtitles}} \\ \hline
\multicolumn{1}{l}{{\sc Entertainment}} &  &  \\
Jokes & --- & \cite{russian-jokes-dataset} \\ \hline
\multicolumn{1}{l}{{\sc Finance}} & \cellcolor[HTML]{EFEFEF}\cite{finance-fr} & \cellcolor[HTML]{EFEFEF}ruscorpora.ru \\ \hline
\multicolumn{1}{l}{{\sc Speech}} & \multicolumn{1}{c}{} & \multicolumn{1}{c}{} \\
Ted Talks & ted.com/talks & ted.com/talks \\
Public Speech & \cellcolor[HTML]{EFEFEF}--- & \cellcolor[HTML]{EFEFEF}ruscorpora.ru \\ \hline
\multicolumn{1}{l}{{\sc Statements}} &  &  \\
Quotes & evene.lefigaro.fr & infoselection.ru \\ \hline
\multicolumn{1}{l}{{\sc Poetry}} & \cellcolor[HTML]{EFEFEF}poesie-francaise.fr & \cellcolor[HTML]{EFEFEF}ruscorpora.ru \\ \hline
\multicolumn{1}{l}{{\sc Letters}} & gutenberg.org & runivers.ru \\ \hline
\end{tabular}
\caption{Dataset Sources (1/2). (---)  denotes that no resource was found in the particular language.}
\end{table*}

\newpage

\newpage

\newpage

\begin{table*}[]
\centering
\begin{adjustbox}{width=0.9\linewidth}
\begin{tabular}{rlll}
\hline
\multicolumn{1}{l}{\textbf{Domain}} & \textbf{Source} & \textbf{Type} & \textbf{License} \\ \hline
\textbf{Sub-Domain} & \textbf{} & \textbf{} & \textbf{} \\ \hline
\multicolumn{1}{l}{{\sc Wikipedia}} & wikipedia.com & Web Article & CC BY-SA 3.0 \\ \hline
\multicolumn{1}{l}{\multirow{2}{*}{{\sc News Articles}}} & \cite{english-news} & Public Dataset & CC BY 4.0 \\
\multicolumn{1}{l}{} & \cite{arabic-news} & Public Dataset & CC BY 4.0 \\ \hline
\multicolumn{4}{l}{{\sc Research}} \\
\multirow{3}{*}{Law} & spu.sharjah.ac.ae & Research Article & CC BY 4.0 \\
 & elgaronline.com & Research Article & CC BY 4.0 \\
 & library.bjp.org & Research Article & CC \\ \cline{2-4} 
\multirow{3}{*}{Politics} & jcopolicy.uobaghdad.edu.iq & Research Article & CC BY 4.0 \\
 & tandfonline.com & Research Article & CC BY 4.0 \\
 & journal.ijarms.org & Research Article & CC \\ \cline{2-4} 
Medical & onlinelibrary.wiley.com & Research Article & CC BY-NC \\ \cline{2-4} 
Literature & jstor.org/journal/jmodelite & Research Article & CC \\
\multicolumn{1}{l}{} & hindijournal.com & Research Article & CC \\ \cline{2-4} 
\multirow{3}{*}{Economics} & asjp.cerist.dz/index.php/en & Research Article & CC \\
 & aeaweb.org & Research Article & CC BY 4.0 \\
 & journal.ijarms.org & Research Article & CC BY 4.0 \\ \hline
\multirow{3}{*}{Science \& Engineering} & arxiv.org & Research Article & CC BY 4.0 \\
 & hal.science & Research Article & CC \\
 & ruscorpora.ru & Research Article & Personal/Non-Commercial \\ \hline
\multicolumn{1}{l}{{\sc Literature}} & hindawi.org/books/ & Book & Public Domain \\
\multicolumn{1}{l}{} & gutenberg.org & Book & Public Domain \\ \hline
\multicolumn{1}{l}{\multirow{3}{*}{{\sc Textbooks}}} & hindawi.org/books/ & Book & Public Domain \\
\multicolumn{1}{l}{} & open.umn.edu & Book & CC BY 4.0 \\
\multicolumn{1}{l}{} & ncert.nic.in & Book & Public Domain \\ \hline
\multicolumn{1}{l}{{\sc Legal}} &  &  &  \\
Constitutions & presidency.gov.lb & Document & Public Domain \\
\multicolumn{1}{l}{} & constitutioncenter.org & Document & CC BY-NC-ND 4.0 \\
\multicolumn{1}{l}{} & legifrance.gouv.fr & Document & Public Domain \\
\multicolumn{1}{l}{} & legislative.gov.in & Document & Public Domain \\
\multicolumn{1}{l}{} & constitution.ru & Document & Public Domain \\ \cline{2-4} 
\multirow{2}{*}{Judicial Rulings} & law.cornell.edu/supremecourt & Document & CC BY-NC-SA 2.5 \\
 & HLDC \cite{judicial-hi} & Public Dataset & Public Domain \\
\multicolumn{1}{l}{} & supcourt.ru & Document & Public Domain \\
UN Parliament & UN Parallel Corpus \cite{un-parallel-corpus} & Public Dataset & Public Domain \\ \hline
\end{tabular}
\end{adjustbox}
\caption{License or term of use per source (1/3)}
\label{tab:license-1}
\end{table*}

\newpage

\begin{table*}[]
\centering
\begin{adjustbox}{width=\linewidth}    
\begin{tabular}{@{}rlll@{}}
\toprule
\multicolumn{1}{l}{\textbf{Domain}} & \textbf{Source} & \textbf{Type} & \textbf{License} \\ \midrule
\textbf{Sub-Domain} & \textbf{} & \textbf{} & \textbf{} \\ \midrule
\multicolumn{1}{l}{{\sc User Reviews}} &  &  &  \\
\multirow{3}{*}{Products} & \cite{reviews-ar} & Public Dataset & Public Domain \\
 & MARC \cite{product-review-en} & Public Dataset & Public Domain \\
 & \cite{review-product-hi} & On Request Dataset & --- \\
\multicolumn{1}{l}{} & RuReviews \cite{ru-product-review} & Public Dataset & Apache-2.0 License \\ \cmidrule(l){2-4} 
Books & LABR \cite{book-review-ar} & Public Dataset & GPL-2.0 \\
 & \cite{good-reads} & Public Dataset & Public Domain \\ \cmidrule(l){2-4} 
\multirow{2}{*}{Movies} & JMURv1 \cite{movie-review-en} & Public Dataset & Public Domain \\
 & \cite{hindi-movie-reviews} & Public Dataset & CC BY-SA 4.0 \\ \cmidrule(l){2-4} 
\multirow{2}{*}{Hotels} & \cite{reviews-ar} & Public Dataset & Public Domain \\
 & \cite{hotel-reviews-en} & Public Dataset & CC BY 4.0 \\ \cmidrule(l){2-4} 
\multirow{2}{*}{Restaurants} & \cite{reviews-ar} & Public Dataset & Public Domain \\
 & \cite{trip-advisor-dataset} & Public Dataset & Apache 2.0 \\ \midrule
\multicolumn{1}{l}{{\sc Dialogue}} &  &  &  \\
Open-domain & ArabicED \cite{open-domain-ar} & Public Dataset & MIT License \\
\multicolumn{1}{l}{} & DailyDialog \cite{open-domain-en} & Public Dataset & CC BY-NC-SA 4.0 \\
\multicolumn{1}{l}{} & MDIA \cite{open-domain-hi} & Public Dataset & CC BY 4.0 \\ \cmidrule(l){2-4} 
Negotiation & CraigslistBargain \cite{negotiation-en} & Public Dataset & MIT License \\ \cmidrule(l){2-4} 
Task-oriented & xSID \cite{task-oriented-en} & Public Dataset & CC BY 4.0 \\
\multicolumn{1}{l}{} & M-CID \cite{task-oriented-fr} & Public Dataset & Public Domain \\
 & HDRS \cite{task-oriented-hi} & Public Dataset & CC BY-NC 4.0 \\ \midrule
\multicolumn{1}{l}{{\sc Finance}} & \cite{finance} & Public Dataset & CC BY-NC-SA 3.0 \\
\multicolumn{1}{l}{} & CoFiF \cite{finance-fr} & Public Dataset & CC BY-NC 4.0 \\
\multicolumn{1}{l}{} & ruscorpora.ru & Document & Personal/Non-Commercial \\ \midrule
\multicolumn{1}{l}{{\sc Forums}} &  &  &  \\
Reddit & files.pushshift.io/reddit & User Posts & Public Domain \\ \cmidrule(l){2-4} 
QA Websites & CQA-MD \cite{qa-ar} & Public Dataset & Public Domain \\
\multicolumn{1}{l}{} & quora.com \cite{qa-en} & Public Dataset & Public Domain \\
\multicolumn{1}{l}{} & FQuAD \cite{qa-fr} & Public Dataset & Personal/Non-Commercial \\
\multicolumn{1}{l}{} & \cite{hindi-qa} & Public Dataset & Public Domain \\
 & SberQuAD \cite{qa-ru} & Public Dataset & Apache-2.0 License \\ \cmidrule(l){2-4} 
StackOverflow & \cite{stackoverflow} & Public Dataset & MIT License \\ \midrule
\multicolumn{4}{l}{{\sc Social Media}} \\
Twitter & Stanceosaurus \cite{stanceosaurus} & Public Dataset & Developer Agreement and Policy \\
\multicolumn{1}{l}{} & \cite{twitter-fr} & Public Dataset & CC BY-NC 4.0 \\
\multicolumn{1}{l}{} & RuSentiTweet \cite{twitter-ru} & Public Dataset & Public Domain \\ \midrule
\multicolumn{1}{l}{{\sc Policies}} &  &  &  \\
Contracts & ejar.sa / hud.gov & Document & Public Domain \\
\multicolumn{1}{l}{} & cesu.urssaf.fr & Document & Public Domain \\
\multicolumn{1}{l}{} & blanker.ru & Document & Public Domain \\
\multicolumn{1}{l}{} & honeybook.com & Document & Public Domain \\ \cmidrule(l){2-4} 
Olympic Rules & resources.specialolympics.org & Document & Personal/Non-Commercial \\ \cmidrule(l){2-4} 
\multirow{2}{*}{Code of Conduct} & fatimafellowship.com & Web Article & Personal/Non-Commercial \\
 & lonza.com & Document & Personal/Non-Commercial \\ \midrule
\multicolumn{1}{l}{{\sc Guides}} &  &  &  \\
\multirow{2}{*}{User Manuals} & samsung.com/us/support/downloads & Document & Personal/Non-Commercial \\
 & manuals.plus/ru & Web Article & Personal/Non-Commercial \\ \cmidrule(l){2-4} 
Online Tutorials & wikihow.com & Web Article & CC BY-NC-SA 3.0 \\ \cmidrule(l){2-4} 
\multirow{2}{*}{Cooking Recipes} & wikibooks.org & Web Article & CC BY-SA 3.0 \\
 & narendramodi.in & Web Article & Personal/Non-Commercial \\
Code Documentation & mathworks.com & Documentation & Personal/Non-Commercial \\ \bottomrule
\end{tabular}
\end{adjustbox}
\caption{License or term of use per source (2/3)}
\label{tab:license-2}
\end{table*}

\newpage

\begin{table*}[]
\centering
\begin{adjustbox}{width=0.85\linewidth}    
\begin{tabular}{@{}llll@{}}
\toprule
\textbf{Domain} & \textbf{Source} & \textbf{Type} & \textbf{License} \\ \midrule
\multicolumn{1}{r}{\textbf{Sub-Domain}} & \textbf{} & \textbf{} & \textbf{} \\ \midrule
\multicolumn{4}{l}{{\sc Captions}} \\
\multicolumn{1}{r}{Images} & \cite{image-captions-ar} & Public Dataset & Public Domain \\
 & Flikr30K \cite{image-captions-en} & Public Dataset & CC0 \\
 & WikiCaps \cite{image-captions-fr-ru} & Public Dataset & CC BY 4.0 \\
 & \cite{image-captions-hi} & Public Dataset & Public Domain \\
\multicolumn{1}{r}{Videos} & Vatex \cite{video-captions-en} & Public Dataset & CC BY 4.0 \\
 & MultiCapCLIP \cite{video-captions-fr} & Public Dataset & BSD-3-Clause license \\
 & \cite{video-captions-hi} & Public Dataset & Public Domain \\
\multicolumn{1}{r}{Movies} & OpenSubtitles2016 \cite{open-subtitles} & Public Dataset & Public Domain \\
\multicolumn{1}{r}{YouTube} & youtube.com & Captions & CC \\ \midrule
{\sc Medical Text} &  &  &  \\
\multicolumn{1}{r}{Clinical Reports} & i2b2/VA \cite{clinical-reports-en} & On Request Dataset & --- \\ \midrule
\multicolumn{4}{l}{{\sc Dictionaries}} \\
\multicolumn{1}{r}{} & almaany.com & Web Article & CC \\
 & dictionary.com & Web Article & CC \\ \midrule
\multicolumn{4}{l}{{\sc Entertainment}} \\
\multicolumn{1}{r}{\multirow{4}{*}{Jokes}} & \cite{arabic-humour} & Public Dataset & Public Domain \\
\multicolumn{1}{r}{} & \cite{jokes-en} & Public Dataset & MIT License \\
\multicolumn{1}{r}{} & \cite{russian-jokes-dataset} & Public Dataset & Public Domain \\
\multicolumn{1}{r}{} & 123hindijokes.com & Web List & Public Domain \\ \midrule
\multicolumn{4}{l}{{\sc Speech}} \\
\multicolumn{1}{r}{Ted Talks} & ted.com/talks & Video Transcription & CC BY-NC-ND 4.0 \\ \cmidrule(l){2-4} 
\multicolumn{1}{r}{\multirow{3}{*}{Public Speech}} & state.gov/translations/arabic & Web Article & Public Domain \\
\multicolumn{1}{r}{} & ruscorpora.ru & Document & Personal/Non-Commercial \\
\multicolumn{1}{r}{} & whitehouse.gov & Web Article & CC BY 3.0 US \\ \midrule
\multicolumn{4}{l}{{\sc Statements}} \\
\multicolumn{1}{r}{Rumours} & Stanceosaurus \cite{stanceosaurus} & Public Dataset & Public Domain \\ \cmidrule(l){2-4} 
\multicolumn{1}{r}{\multirow{5}{*}{Quotes}} & arabic-quotes.com & Web List & Public Domain \\
\multicolumn{1}{r}{} & goodreads.com/quotes & Web List & Public Domain \\
\multicolumn{1}{r}{} & evene.lefigaro.fr & Web List & Personal/Non-Commercial \\
\multicolumn{1}{r}{} & storyshala.in & Web List & Public Domain \\
\multicolumn{1}{r}{} & infoselection.ru & Web List & Personal/Non-Commercial \\ \midrule
\multirow{5}{*}{{\sc Poetry}} & aldiwan.net & Web List & Public Domain \\
 & poetryfoundation.org & Web List & Public Domain \\
 & poesie-francaise.fr & Web List & Public Domain \\
 & hindionlinejankari.com & Web List & Public Domain \\
 & ruscorpora.ru & Document & Personal/Non-Commercial \\ \midrule
{\sc Letters} & oflosttime.com & Web Article & Public Domain \\
 & gutenberg.org & Document & Public Domain \\
 & runivers.ru & Document & Personal/Non-Commercial \\ \bottomrule
\end{tabular}
\end{adjustbox}
\caption{License or term of use per source (3/3)}
\label{tab:license-3}
\end{table*}

\end{document}